\pdfoutput=1
\documentclass[10pt,twocolumn,letterpaper]{article}

\usepackage{iccv}
\usepackage{times}
\usepackage{epsfig}
\usepackage{graphicx}
\usepackage{amsmath}
\usepackage{amssymb}
\usepackage{algorithm}
\usepackage{algpseudocode}
\usepackage{boldline}
\usepackage{tabularx}

\usepackage[breaklinks=true,bookmarks=false]{hyperref}

\iccvfinalcopy 

\def\iccvPaperID{****} 

\DeclareMathOperator*{\argmin}{argmin}
\DeclareMathOperator*{\argmax}{argmax}

\begin{document}

\title{Image Restoration using Autoencoding Priors}

\author{Siavash Arjomand Bigdeli$^1$\\
$^1$University of Bern \\
{\tt\small bigdeli@inf.unibe.ch}
\and
Matthias Zwicker$^{1,2}$\\
$^2$University of Maryland, College Park\\
{\tt\small zwicker@inf.unibe.ch}
}

\maketitle

\begin{abstract}
We propose to leverage denoising autoencoder networks as priors to address image restoration problems. We build on the key observation that the output of an optimal denoising autoencoder is a local mean of the true data density, and the autoencoder error (the difference between the output and input of the trained autoencoder) is a mean shift vector. We use the magnitude of this mean shift vector, that is, the distance to the local mean, as the negative log likelihood of our natural image prior. For image restoration, we maximize the likelihood using gradient descent by backpropagating the autoencoder error. A key advantage of our approach is that we do not need to train separate networks for different image restoration tasks, such as non-blind deconvolution with different kernels, or super-resolution at different magnification factors. We demonstrate state of the art results for non-blind deconvolution and super-resolution using the same autoencoding prior.
\end{abstract}

\section{Introduction}

\begin{figure}[t]
\centering
\begin{tabular}{ccccc}
Blurry & \multicolumn{4}{c}{ Iterations 10, 30, 100 and 250} \\
$23.12dB$ & $24.17dB$ & $26.43dB$ & $29.1dB$ & $29.9dB$
\end{tabular}

\includegraphics[width=.19\columnwidth]{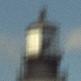} 
\includegraphics[width=.19\columnwidth]{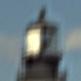}
\includegraphics[width=.19\columnwidth]{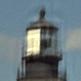}
\includegraphics[width=.19\columnwidth]{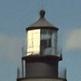}
\includegraphics[width=.19\columnwidth]{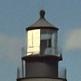} \\
\vspace{1pt}
\includegraphics[width=.19\columnwidth]{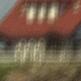} 
\includegraphics[width=.19\columnwidth]{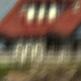}
\includegraphics[width=.19\columnwidth]{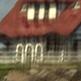}
\includegraphics[width=.19\columnwidth]{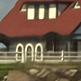}
\includegraphics[width=.19\columnwidth]{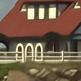} 

\caption{
We propose a natural image prior based on a denoising autoencoder, and apply it to image restoration problems like non-blind deblurring. The output of an optimal denoising autoencoder is a local mean of the true natural image density, and the autoencoder error is a mean shift vector. We use the magnitude of the mean shift vector as the negative log likelihood of our prior. To restore an image from a known degradation, we use gradient descent to iteratively minimize the mean shift magnitude while respecting a data term. Hence, step-by-step we shift our solution closer to its local mean in the natural image distribution.
}
\label{fig:teaser}
\end{figure}

Deep learning has been successful recently at advancing the state of the art in various low-level image restoration problems including image super-resolution, deblurring, and denoising. The common approach to solve these problems is to train a network end-to-end for a specific task, that is, different networks need to be trained for each noise level in denoising, or each magnification factor in super-resolution. This makes it hard to apply these techniques to related problems such as non-blind deconvolution, where training a network for each blur kernel would be impractical.

A standard strategy to approach image restoration problems is to design suitable priors that can successfully constrain these underdetermined problems. Classical techniques include priors based on edge statistics, total variation, sparse representations, or patch-based priors. In contrast, our key idea is to leverage denoising autoencoder (DAE) networks~\cite{Vincent:2008:ECR} as natural image priors. 
We build on the key observation by Alain et al.~\cite{JMLR:v15:alain14a} that for each input, the output of an optimal denoising autoencoder is a local mean of the true natural image density. The weight function that defines the local mean is equivalent to the noise distribution used to train the DAE. In other words, the autoencoder error, which is the difference between the output and input of the trained autoencoder, is a mean shift vector~\cite{Comaniciu:2002:MSR}, and the noise distribution represents a mean shift kernel. 

Hence, we leverage neural DAEs in an elegant manner to define powerful image priors: Given the trained autoencoder, our natural image prior is based on the magnitude of the mean shift vector. For each image, the mean shift is proportional to the gradient of the true data distribution smoothed by the mean shift kernel, and its magnitude is the distance to the local mean in the distribution of natural images. With an optimal DAE, the energy of our prior vanishes exactly at the stationary points of the true data distribution smoothed by the mean shift kernel. This makes our prior attractive for maximum a posteriori (MAP) estimation.

For image restoration, we include a data term based on the known image degradation model. For each degraded input image, we maximize the likelihood of our solution using gradient descent by backpropagating the autoencoder error and computing the gradient of the data term. Intuitively, this means that our approach iteratively moves our solution closer to its local mean in the natural image density, while satisfying the data term. This is illustrated in Figure~\ref{fig:teaser}.

A key advantage of our approach is that we do not need to train separate networks for different image restoration tasks, such as non-blind deconvolution with different kernels, or super-resolution at different magnification factors. Even though our autoencoding prior is trained on a denoising problem, it is highly effective at removing these different degradations. We demonstrate state of the art results for non-blind deconvolution and super-resolution using the same autoencoding prior.

\section{Related Work}

Image restoration, including deblurring, denoising, and super-resolution, is an underdetermined problem that needs to be constrained by effective priors to obtain acceptable solutions. Without attempting to give a complete list of all relevant contributions, the most common successful techniques include priors based on edge statistics~\cite{Fattal:2007:IUV,Tappen:03:ESD}, total variation~\cite{Perrone:2014:TVB}, sparse representations~\cite{Aharon:2006:KSVD,Yang:2010:ISR}, and patch-based priors~\cite{Zoran:2011:LMN,Levin:2012:PCF,Schmidt:2016:CRT}. While some of these techniques are tailored for specific restoration problems, recent patch-based priors lead to state of the art results for multiple applications, such as deblurring and denoising~\cite{Schmidt:2016:CRT}.

Solving image restoration problems using neural networks seems attractive because they allow for straightforward end-to-end learning. This has led to remarkable success for example for single image super-resolution~\cite{Dong:2014:DSR,Gu:2015:CSC,Dong:2016:ISR,Liu:2016:RSI,Kim:2016:AIS} and denoising~\cite{Burger:2012:IDC,Mao:2016:IRV}. A disadvantage of the end-to-end learning is that, in principle, it requires training a different network for each restoration task (e.g., each different noise level or magnification factor). While a single network can be effective for denoising different noise levels~\cite{Mao:2016:IRV}, and similarly a single network can perform well for different super-resolution factors~\cite{Kim:2016:AIS}, 
it seems unlikely that in non-blind deblurring, the same network would work well for arbitrary blur kernels.
Additionally, experiments by Zhang et al.~\cite{zhang2016beyond} show that training a network for multiple tasks reduces performance compared to training each task on a separate network.
Previous research addressing non-blind deconvolution using deep networks includes the work by Schuler et al.~\cite{Schuler:2013:MLA} and more recently Xu et al.~\cite{Xu:2014:DCN}, but they require end-to-end training for each blur kernel.

A key idea of our work is to train a neural autoencoder that we use as a prior for image restoration. Autoencoders are typically used for unsupervised representation learning~\cite{Vincent:2010:SDA}. The focus of these techniques lies on the descriptive strength of the learned representation, which can be used to address classification problems for example. In addition, generative models such as generative adversarial networks~\cite{Goodfellow:2014:GAN} or variational autoencoders~\cite{Kingma:2014:AEV} also facilitate sampling the representation to generate new data. Their network architectures usually consist of an encoder followed by a decoder, with a bottleneck that is interpreted as the data representation in the middle. The ability of autoencoders and generative models to create images from abstract representations makes them attractive for restoration problems. Notably, the encoder-decoder architecture in Mao et al.'s image restoration work~\cite{Mao:2016:IRV} is highly reminiscent of autoencoder architectures, although they train their network in a supervised manner. 

A denoising autoencoder~\cite{Vincent:2008:ECR} is an autoencoder trained to reconstruct data that was corrupted with noise.
Previously, Alain and Bengio~\cite{JMLR:v15:alain14a} and Nguyen et al.~\cite{nguyen2016plug} used DAEs to construct generative models.
We are inspired by the insight of Alain and Bengio that the output of an optimal DAE is a local mean of the true data density. Hence, the autoencoder error (the difference between its output and input) is a mean shift vector~\cite{Comaniciu:2002:MSR}. 
This motivates using the magnitude of the autoencoder error as our prior. 

Our work has an interesting connection to the plug-and-play priors introduced by Venkatakrishnan et al.~\cite{venkatakrishnan2013plug}. They solve regularized inverse (image restoration) problems using ADMM (alternating directions method of multipliers), and they make the key observation that the optimization step involving the prior is a denoising problem, that can be solved with any standard denoiser. Brifman et al.~\cite{brifman2016turning} leverage this framework to perform super-resolution, and they use the NCSR denoiser~\cite{dong2013nonlocally} based on sparse representations. While their use of a denoiser is a consequence of ADMM, our DAE prior is motivated by its relation to the underlying data density (the distribution of natural images). 
Our approach leads to a different, simpler gradient descent optimization that does not rely on ADMM.

In summary, the main contribution of our work is that we show how to leverage DAEs to define a prior for image restoration problems by making the connection to mean shift. Crucially, for each image our prior is the squared distance to its local mean in the natural image distribution. We train a DAE and demonstrate that the resulting prior is effective for different restoration problems, including deblurring with arbitrary kernels and super-resolution with different magnification factors.

\section{Problem Formulation}

We formulate image restoration in a standard fashion as a maximum a posteriori (MAP) problem~\cite{Joshi:2009:IDD}. We model degradation including blur, noise, and downsampling as
\begin{align}
B = D(I \otimes K) + \xi,
\end{align}
where $B$ is the degraded image, $D$ is a down-sampling operator using point sampling, $I$ is the unknown image to be recovered, $K$ is a known, shift-invariant blur kernel, and $\xi \sim \mathcal{N}(0,\sigma_d^2)$ is the per-pixel i.i.d. degradation noise. The posterior probability of the unknown image is $p(I|B) = p(B|I)p(I)/p(B)$, and we maximize it by minimizing the corresponding negative log likelihoods $L$,
\begin{align}
\argmax_I p(I|B) = \argmin_I \left[ L(B|I) + L(I) \right].
\end{align}
Under the Gaussian noise model, the negative data log likelihood is
\begin{align}
L(B|I) = \|B-D(I \otimes K)\|^2 / \sigma_d^2.
\end{align}
Note that this implies that the blur kernel $K$ is given at the higher resolution, before down-sampling by point sampling with $D$. Our contribution now lies in a novel image prior $L(I)$, which we introduce next.

\section{Denoising Autoencoder as Natural Image Prior}

\begin{figure}[t]
\centering
\small

\begin{tabular}{cc}
(a) Spiral Manifold& (b) Smoothed Density\\
and Observed Samples & from Observed Samples \\
\includegraphics[width=.45\columnwidth]{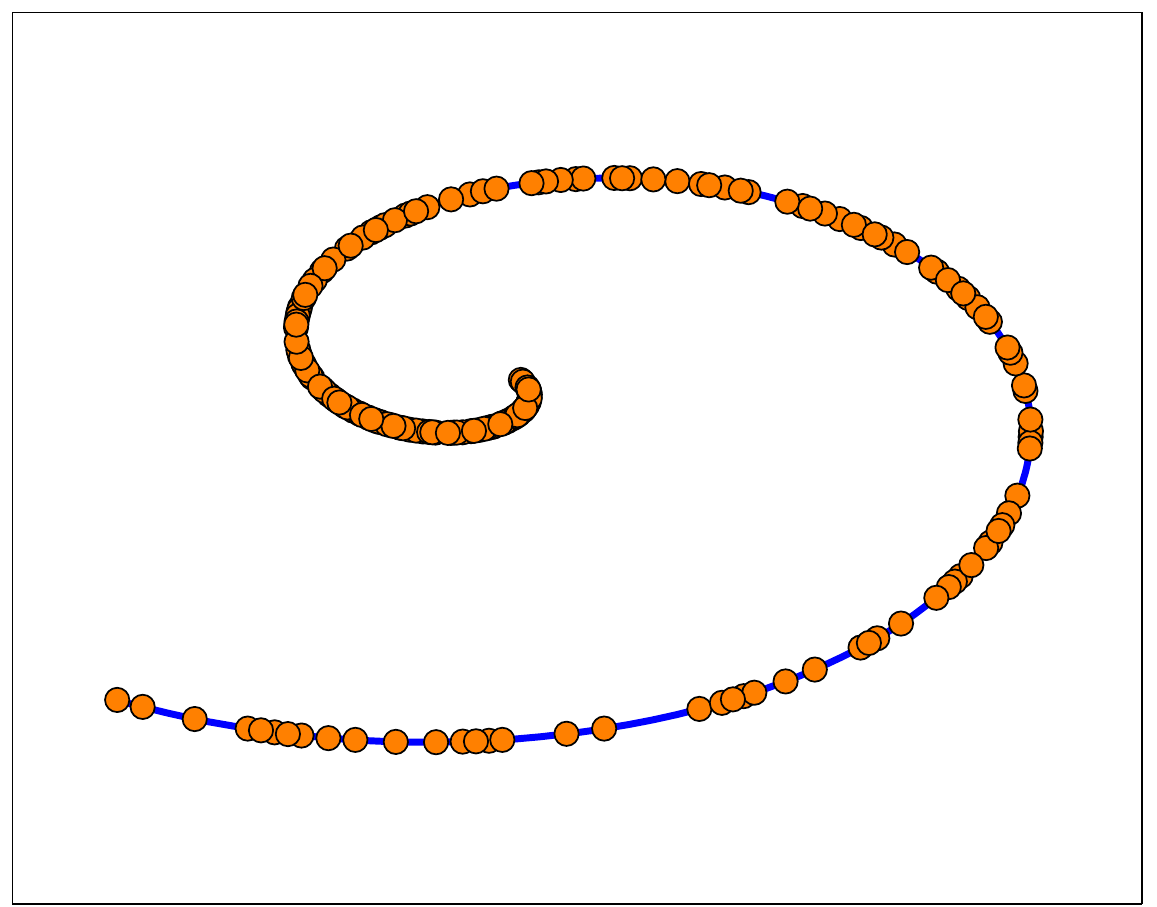}&
\includegraphics[width=.45\columnwidth]{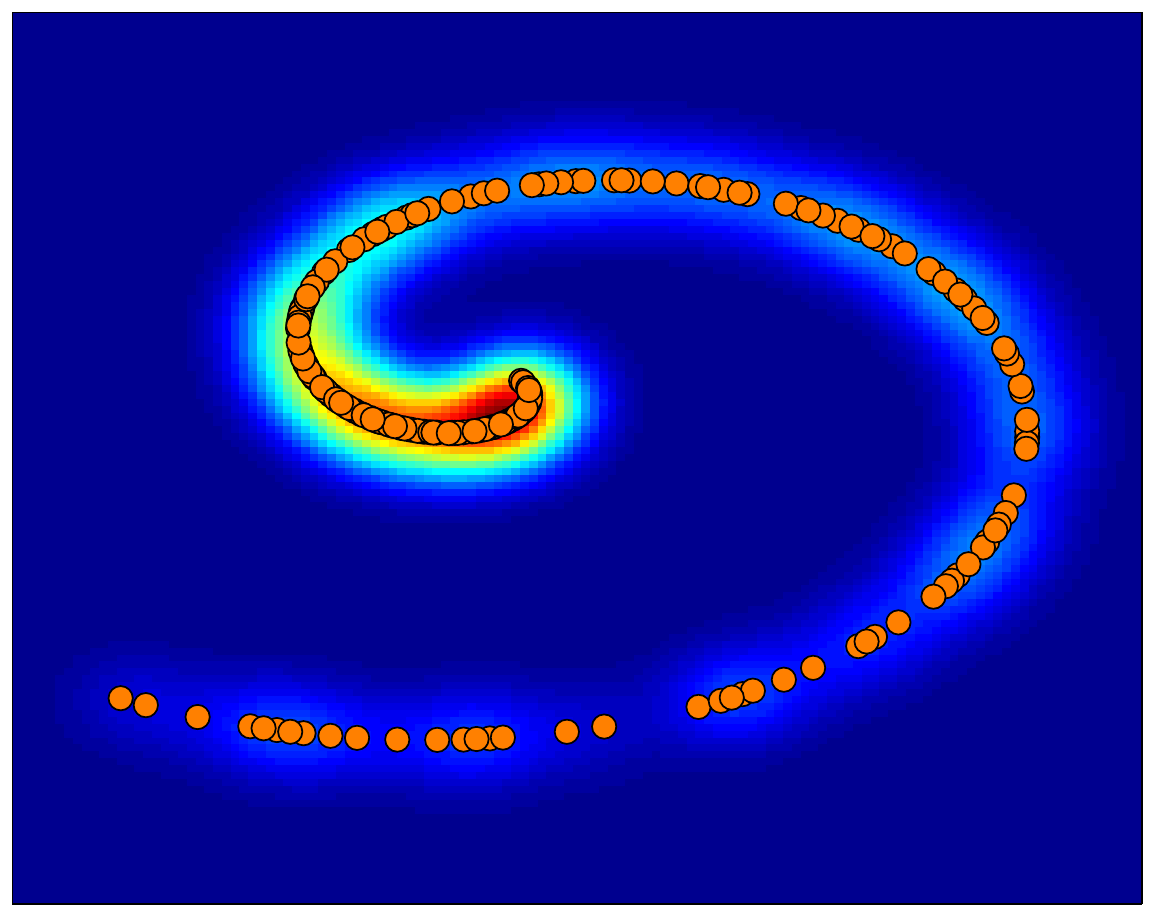} \\
(c) Mean Shift Vectors & (d) Mean Shift Vectors\\
Learned by DAE  & Approximated (Eq.~\ref{eq:meanShiftR})\\
\includegraphics[width=.45\columnwidth]{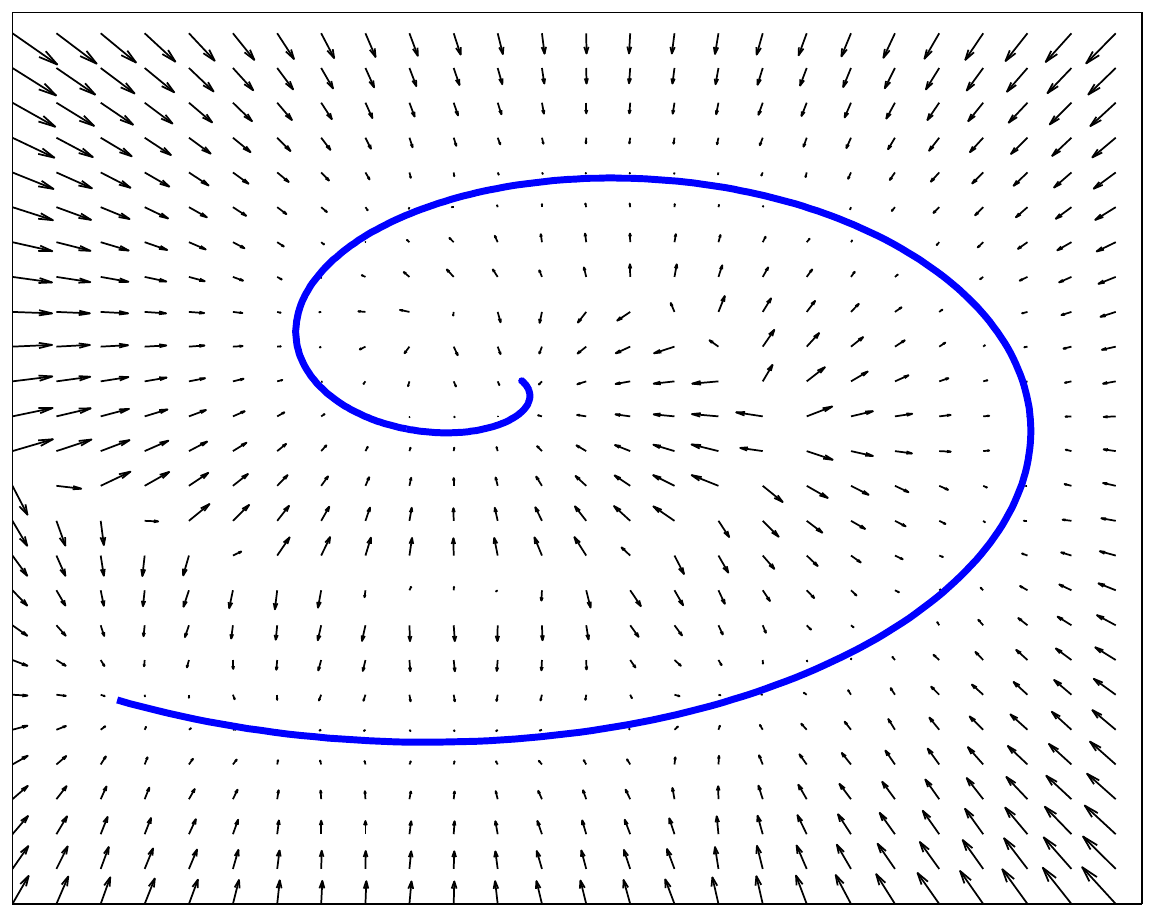}&
\includegraphics[width=.45\columnwidth]{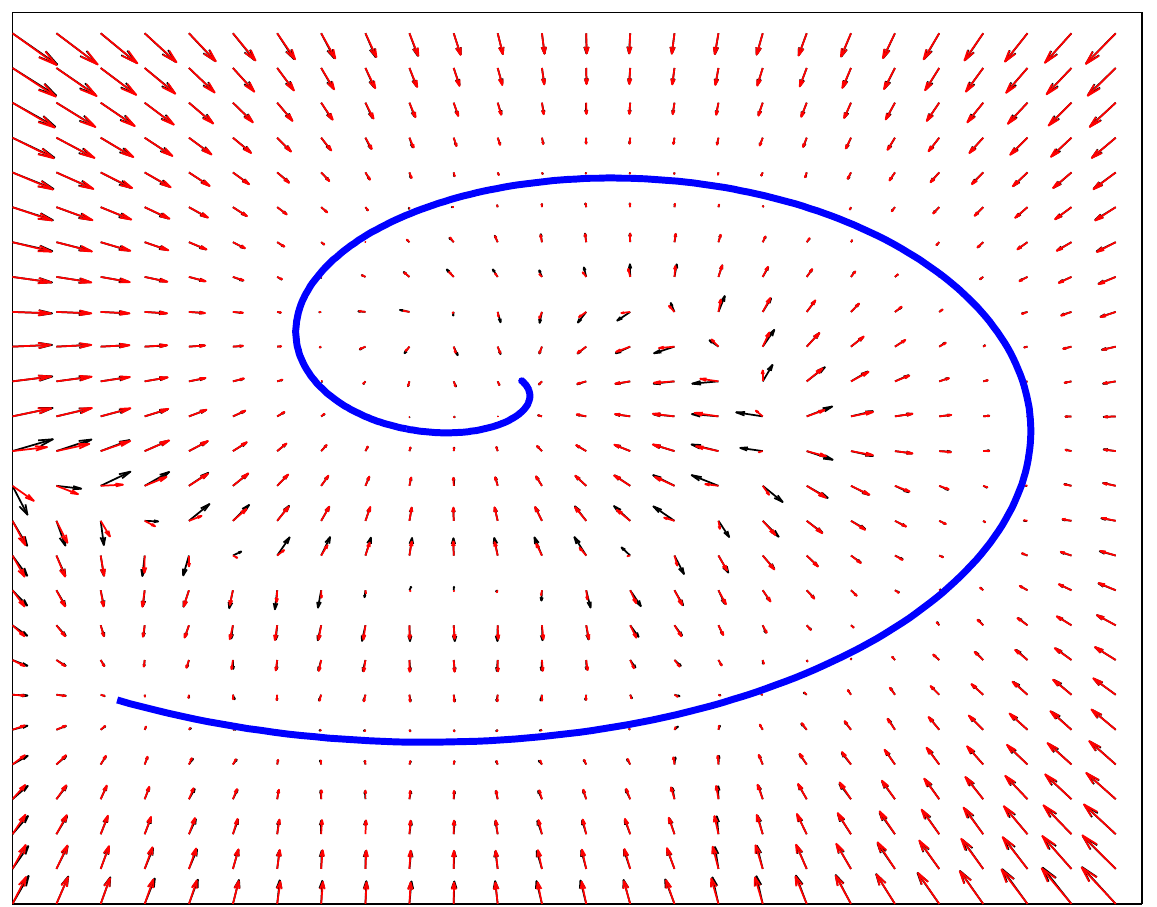} 
\end{tabular}

\caption
{
Visualization of a denoising autoencoder using a 2D spiral density. Given input samples of a true density (a), the autoencoder is trained to pull each sample corrupted by noise back to its original location. Adding noise to the input samples smooths the density represented by the samples (b). Assuming an infinite number of input samples and an autoencoder with unlimited capacity, for each input, the output of the optimal trained autoencoder is the local mean of the true density. The local weighting function corresponds to the noise distribution that was used during training, and it represents a mean shift kernel~\cite{Comaniciu:2002:MSR}. The difference between the output and the input of the autoencoder is a mean shift vector (c), which vanishes at local extrema of the true density smoothed by the mean shift kernel.
Due to practical limitations (Section~\ref{sec:AEPrior}), we approximate the mean shift vectors (d, red) using Equation~\ref{eq:meanShiftR}. The difference between the true mean shift vectors (d, black) and our approximate vectors (d, red) vanishes as we get closer to the manifold.
}
\label{fig:densityPlot}
\end{figure}

%

Our key idea is to leverage a neural autoencoder to define a natural image prior. In particular, we are building on denoising autoencoders (DAE)~\cite{Vincent:2008:ECR} that are trained using Gaussian noise and an expected quadratic loss. We are inspired by the results by Alain et al.~\cite{JMLR:v15:alain14a} who show how the output of such autoencoders relates to the underlying data density, and we will exploit this relation to define our prior. 

\subsection{Denoising Autoencoders}

We visualize the intuition behind DAEs in Figure~\ref{fig:densityPlot}. Let us denote a DAE as $A_{\sigma_{\!\eta}}$. Given an input image $I$, its output is an image $A_{\sigma_{\!\eta}}(I)$. A DAE $A_{\sigma_{\!\eta}}$ is trained to minimize~\cite{Vincent:2008:ECR} 
\begin{align}
\mathcal{L}_{\text{DAE}} = \mathbb{E}_{\eta, I} \left[ \|I - A_{\sigma_{\!\eta}}(I + \eta)) \|^2 \right],
\end{align}
where the expectation is over all images $I$ and Gaussian noise $\eta$ with variance $\sigma_{\eta}^2$, and $A_{\sigma_{\!\eta}}$ indicates that the DAE was trained with noise variance $\sigma_{\eta}^2$. It is important to note that the noise variance $\sigma_{\eta}^2$ here is not related to the degradation noise and its variance $\sigma_d^2$, and it is not a parameter to be learned. Instead, it is a user specified parameter whose role becomes clear with the following result by Alain et al.~\cite{JMLR:v15:alain14a}: Let us denote the true data density of natural images as $p(I)$. Alain et al. show that the output $A_{\sigma_{\!\eta}}(I)$ of the optimal DAE (assuming unlimited capacity) is related to the true data density $p(I)$ as
\begin{align}
A_{\sigma_{\!\eta}}(I) &= \frac{\mathbb{E}_{\eta}\left[ p(I-\eta)(I-\eta)\right]}{\mathbb{E}_{\eta}\left[p(I-\eta)\right]} \nonumber \\
&= \frac{\int g_{\sigma_{\eta}^2}(\eta) p(I- \eta) (I- \eta)  d \eta}{\int g_{\sigma_{\eta}^2}(\eta) p(I- \eta) d \eta}.
\end{align}
%
%
%
%
%
This means that the autoencoder output can be interpreted as a local mean or a weighted average of images in the neighborhood of $I$. 
The weights are given by the true density $p(I)$ multiplied by the noise distribution that was used during training, which is a local Gaussian kernel $g_{\sigma_{\eta}^2}(\eta)$ centered at $I$ with variance $\sigma_{\eta}^2$. That is, the parameter $\sigma_{\eta}^2$ of the autoencoder determines the size of the region around $I$ that contributes to the local mean. 
This reveals an interesting connection to the mean shift algorithm~\cite{Comaniciu:2002:MSR}: The autoencoder error, that is the difference between the output and the input of the autoencoder $A_{\sigma_{\!\eta}}(I)-I$ is a mean shift vector. 
When the noise has a Gaussian distribution, it is straightforward to show that this autoencoder error is proportional to the gradient of the log likelihood of the smoothed density,
\begin{align}
A_{\sigma_{\!\eta}}(I) - I &= \sigma_{\eta}^2 \nabla \log \mathbb{E}_{\eta}\left[ p(I-\eta) \right] \nonumber \\
&= \sigma_{\eta}^2 \nabla \log \left[ g_{\sigma_{\!\eta}}*p\right](I),
\end{align}
where $*$ means convolution (see supplemental material for derivation). The autoencoder error vanishes at stationary points, including local extrema, of the true density smoothed by the Gaussian kernel.

\begin{figure}[t]
\centering
\includegraphics[width=.325\columnwidth]{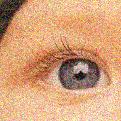} 
\includegraphics[width=.325\columnwidth]{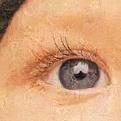}
\includegraphics[width=.325\columnwidth]{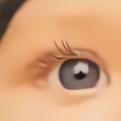} 
\caption
{
Local minimum of our natural image prior. Starting with a noisy image (left), we minimize the prior via gradient descent (middle: intermediate step) to reach the local minimum (right).
}
\label{fig:meanShift}
\end{figure}

\subsection{Autoencoding Prior}
\label{sec:AEPrior}

The above observations inspire us to use the squared magnitude of the mean shift vector as the energy (the negative log likelihood) of our prior, $L(I) = \|A_{\sigma_{\!\eta}}(I)-I\|^2$. This energy is very powerful because it tells us how close an image $I$ is to its local mean $A_{\sigma_{\!\eta}}(I)$ in the true data density, and it vanishes at local extrema of the true density smoothed by the mean shift kernel. Figure~\ref{fig:densityPlot}(c), illustrates how small values of $L(I) = \|A_{\sigma_{\!\eta}}(I)-I\|^2$ occur close to the data manifold, as desired. Figure~\ref{fig:meanShift} visualizes a local minimum of our prior on natural images, which we find by iteratively minimizing the prior via gradient descent starting from a noisy input, without any help from a data term.

Including the data term, we recover latent images as
\begin{align}
\argmin_I \, &\|B-D(I \otimes K)\|^2 / \sigma_d^2 \nonumber \\
& + \gamma  \|A_{\sigma_{\!\eta}}(I)-I\|^2.
\label{eq:lossFunction}
\end{align}
Our energy has two parameters that we will adjust based on the restoration problem. First, this is the mean shift kernel size $\sigma_{\eta}$, and second we introduce a parameter $\gamma$ to weight the relative influence of the data term and the prior. 

\paragraph{Optimization.}
Given a trained autoencoder, we minimize our loss function in Equation~\ref{eq:lossFunction} by applying gradient descent and computing the gradient of the prior using backpropagation through the autoencoder.
Algorithm~\ref{alg:map} shows the steps to minimize Equation~\ref{eq:lossFunction}.
In the first step of each iteration, we compute the gradient of the data term with respect to image $I$.
The second step is to find the gradients for our prior.
The gradient of the mean shift vector $\| A_{\sigma_{\!\eta}}(I) -I \|^2$ requires the gradient of the autoencoder $A_{\sigma_{\!\eta}}(I)$, which we compute by backpropagation through the network.
Finally, the image $I$ is updated using the weighted sum of the two gradient terms.
%
%

\begin{algorithm}[t]
\begin{algorithmic}
\Loop{ $\#iterations$}

\State{$\bullet$}  Compute data term gradients $\nabla_I L(I|B)$:
\begin{center}
$ K^TD^T(DKI-B) / \sigma_d^2 $
\end{center}
\State{$\bullet$}  Compute prior gradients $\nabla_I L(I)$:
\begin{center}
$\nabla_I A_{\sigma_{\!\eta}}(I)^T \Big(A_{\sigma_{\!\eta}}(I)-I\Big) + I - A_{\sigma_{\!\eta}}(I)$
\end{center}
\State{$\bullet$}  Update $I$ by descending
\begin{center}
$\nabla_I L(I|B) + \gamma \nabla_I L(I)$
\end{center}
\EndLoop
\caption{Proposed gradient descent. We express convolution as a matrix-vector product.}
\label{alg:map}
\end{algorithmic}
\end{algorithm}


\paragraph{Overcoming Training Limitations.}
The theory above assumes unlimited data and time to train an unlimited capacity autoencoder.
In particular, to learn the true mean shift mapping, for each natural image the training data needs to include noise patterns that lead to other natural images.
In practice, however, such patterns virtually never occur because of the high dimensionality.
Since the DAE never observed natural images during training (produced by adding noise to other images), it overfits to noisy images.
This is problematic during the gradient descent optimization, when the input to the DAE does not have noise.

As a workaround, we obtained better results by adding noise to the image before feeding it to the trained DAE during optimization.
We further justify this by showing that with this workaround, we can still approximate a DAE that was trained with a desired noise variance $\sigma_{\eta}^2$. 
That is, 
\begin{gather}
\label{eq:meanShiftR}
A_{\sigma_{\!\eta}}(I) - I \approx 2 \Big( \mathbb{E}_{\epsilon} \left[ A_{\sigma_{\epsilon}}(I-\epsilon) \right] - I \Big), 
\end{gather}
%
where $\epsilon \sim \mathcal{N}(0,\sigma_{\epsilon}^2)$, and $A_{\sigma_{\epsilon}}$ is a DAE trained with $\sigma_{\epsilon}^2 = \sigma_{\eta}^2 / 2$. The key point here is that the consecutive convolution with two Gaussians is equivalent to a single Gaussian convolution with the sum of the variances (refer to supplementary material for the derivation).
This is visualized in Figure~\ref{fig:densityPlot}(d). The red vectors indicate the approximated mean shift vectors using Equation~\ref{eq:meanShiftR} and the black vectors indicate the exact mean shift vectors. The approximation error decreases as we approach the true manifold.

During optimization, we approximate the expected value in Equation~\ref{eq:meanShiftR} by stochastically sampling over $\epsilon$. 
We use momentum of $0.9$ and step size $0.1$ in all experiments and we found that using one noise sample per iteration performs well enough to compute meaningful gradients. This approach resulted in a PSNR gain of around $1.7dB$ for the super-resolution task (Section~\ref{sec:sr}), compared to evaluating the left hand side of Equation~\ref{eq:meanShiftR} directly.

\paragraph{Bad Local Minima and Convergence.}
The mean shift vector field learned by the DAE could vanish in low density regions~\cite{JMLR:v15:alain14a}, which corresponds to undesired local minima for our prior. In practice, however, we have not observed such degenerate solutions because our data term pulls the solution towards natural images. In all our experiments the optimization converges smoothly (Figure~\ref{fig:teaser}, intermediate steps), although we cannot give a theoretical guarantee.


\subsection{Autoencoder Architecture and Training}


Our network architecture is inspired by Zhang et al.~\cite{zhang2016beyond}.
The network consists of 20 convolutional layers with batch normalization in between except for the first and last layers, and we use ReLU activations except for the last convolutional layer.
The convolution kernels are of size $3 \times 3$ and the number of channels are 3 (RGB) for input and output and 64 for the rest of the layers. 
Unlike typical neural autoencoders, our network does not have a bottleneck. An explicit latent space implemented as a bottleneck is not required in principle for DAE training, and we do not need it for our application.
We use a fully-convolutional network that allows us to compute the gradients with respect to the image more efficiently since the neuron activations are shared between many pixels.
Our network is trained on color images of the ImageNet dataset~\cite{deng2009imagenet} by adding Gaussian noise with standard deviation $\sigma_{\epsilon} = 25$ (around $10\%$).
We perform residual learning by minimizing the $L_2$ distance of the output layer to the ground truth noise.
We used the \textit{Caffe} package~\cite{jia2014caffe} and employed an Adam solver~\cite{kingma2014adam} with $\beta_1 = 0.9$, $\beta_2 = 0.999$ and learning rate of $0.001$, which we reduced during the iterations.

\section{Experiments and Results}

We compare our approach, Denoising Autoencoder Prior (DAEP), to state of the art methods in super-resolution and non-blind deconvolution problems. For all our experiments, we trained the autoencoder with $\sigma_{\epsilon} = 25$ ($\sigma_{\eta} = 25\sqrt{2}$), and the parameter of our energy (Equation~\ref{eq:lossFunction}) were set to $\gamma = 6.875 / \sigma_{\eta}^2$. We always perform $300$ gradient descent iteration steps during image restoration, which takes about $30$ seconds for a $256 \times 256$ image.

\subsection{Super-Resolution}
\label{sec:sr}

\begin{table}
\begin{center}
\begin{tabularx}{\columnwidth}[c]{lXccccc}

\hlineB{2}
Method && $\times 2$ & $\times 3$ & $\times 4$ & $\times 5$ \\
\hline
Bicubic && 31.80 & 28.67 & 26.73 & 25.32 \\
SRCNN~\cite{Dong:2016:ISR} && 34.50 & 30.84 & 28.60 & 26.12 \\
TNRD~\cite{chen2016trainable} && 34.62 & 31.08 & 28.83 & 26.88 \\
VDSR~\cite{Kim:2016:AIS} && 34.50 & 31.39 & 29.19 & 25.91 \\
DnCNN-3~\cite{zhang2016beyond} && 35.20 & \textbf{31.58} & \textbf{29.30} & 26.30 \\
DAEP (Ours) && \textbf{35.23} & 31.44 & 29.01 & \textbf{27.19} \\
\hlineB{2}
\end{tabularx}
\end{center}

\caption{Average PSNR ($dB$) for super-resolution on 'Set5' ~\cite{Bevilacqua:2012:LCS}. }
\label{tbl:superResolutionSet5}
\end{table}

\begin{table}
\begin{center}
\begin{tabularx}{\columnwidth}[c]{lXccccc}
\hlineB{2}
Method && $\times 2$ & $\times 3$ & $\times 4$ & $\times 5$ \\
\hline
Bicubic && 28.53 & 25.92 & 24.44 & 23.46 \\
SRCNN~\cite{Dong:2016:ISR} && 30.52 & 27.48 & 25.76 & 24.05 \\
TNRD~\cite{chen2016trainable} && 30.53 & 27.60 & 25.92 & 24.61 \\
VDSR~\cite{Kim:2016:AIS} && 30.72 & 27.81 & 26.16 & 24.01 \\
DnCNN-3~\cite{zhang2016beyond} && 30.99 & 27.93 & \textbf{26.25} & 24.26 \\
DAEP (Ours) && \textbf{31.07} & \textbf{27.93} & 26.13 & \textbf{24.88} \\
\hlineB{2}
\end{tabularx}
\end{center}

\caption{Average PSNR ($dB$) for super-resolution on 'Set14'~\cite{Zeyde:2010:SIS}. }
\label{tbl:superResolutionSet14}
\end{table}

\begin{figure*}[t]
\centering
\bgroup
\setlength{\tabcolsep}{5pt}
\begin{tabular}{cccccc}
Ground Truth & Bicubic & SRCNN~\cite{Dong:2016:ISR} & TNRD~\cite{chen2016trainable} & DnCNN-3~\cite{zhang2016beyond} & DAEP (Ours) \\
\includegraphics[width=.145\textwidth]{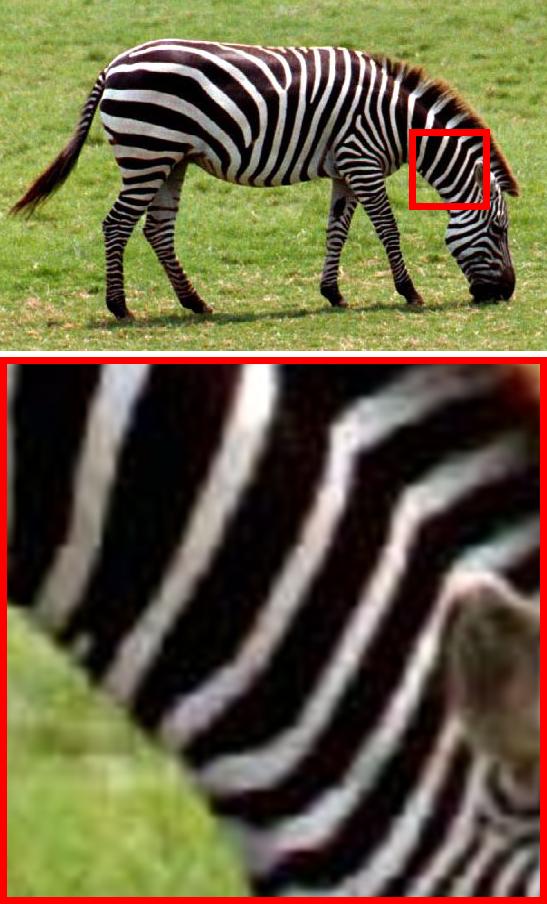} &
\includegraphics[width=.145\textwidth]{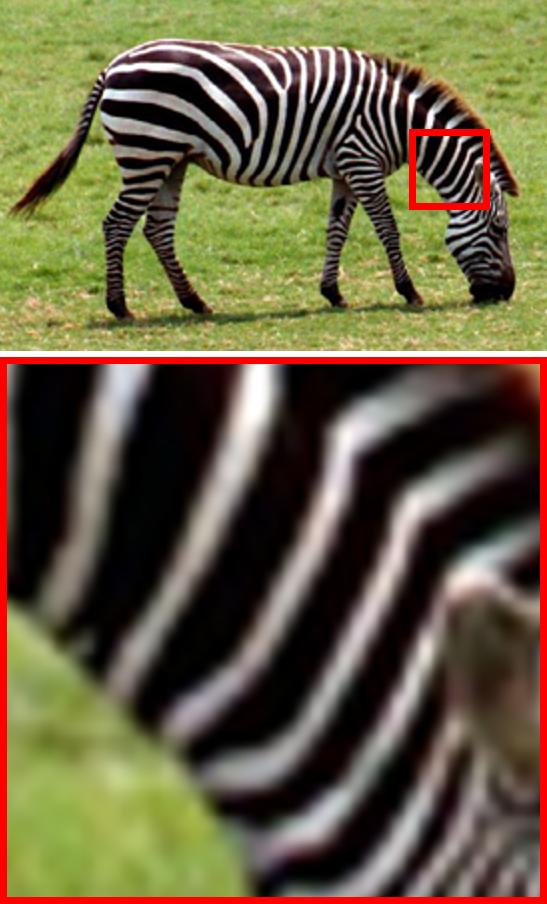} &
\includegraphics[width=.145\textwidth]{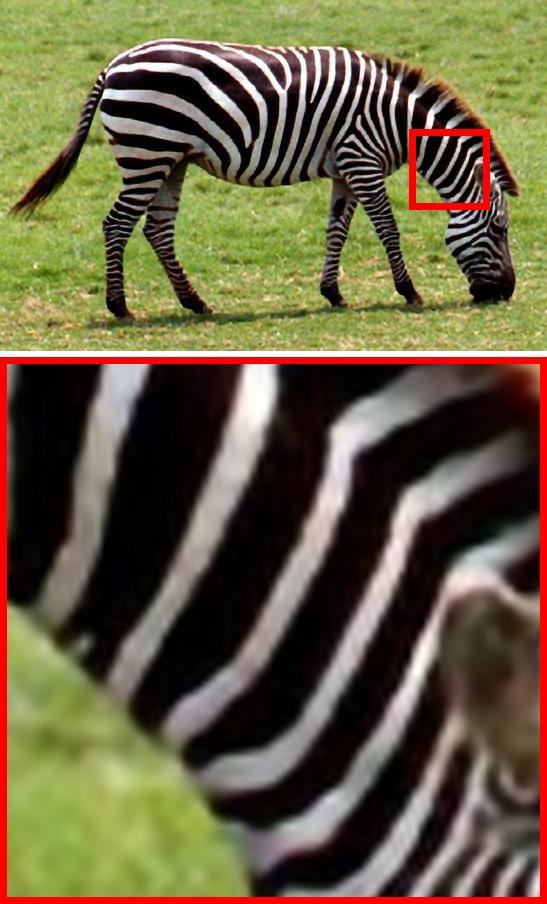} &
\includegraphics[width=.145\textwidth]{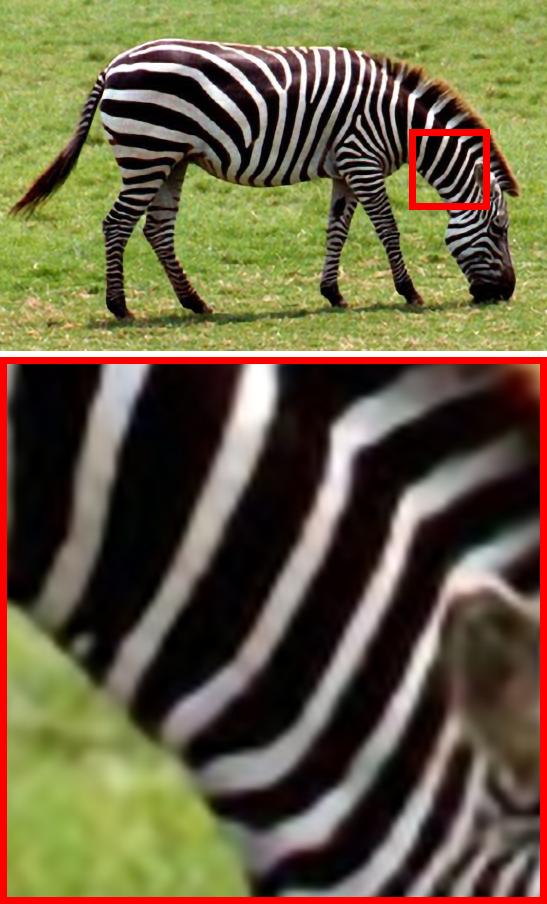} &
\includegraphics[width=.145\textwidth]{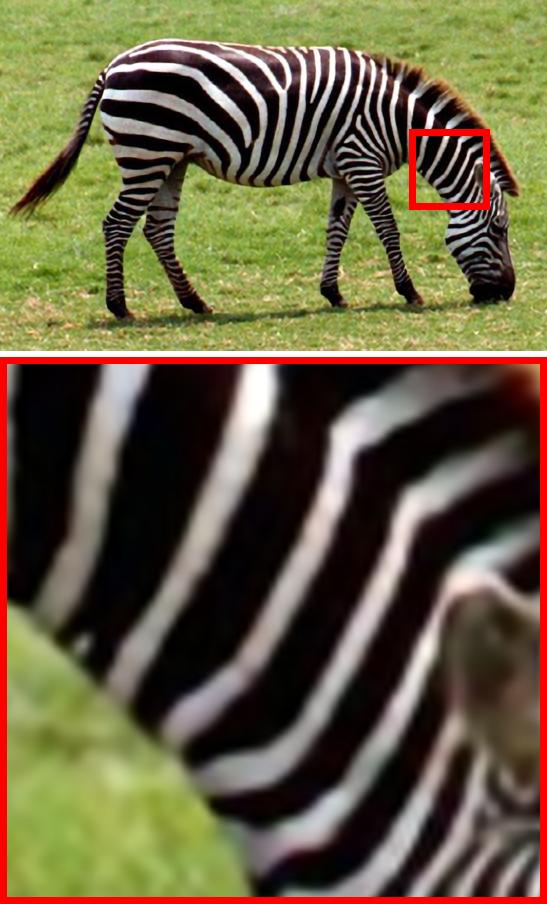} &
\includegraphics[width=.145\textwidth]{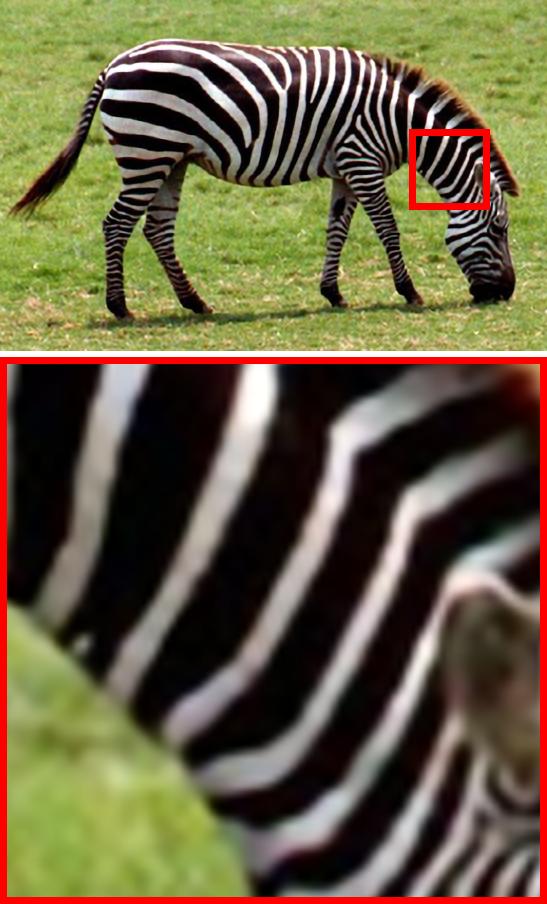} \\
 & $29.12$ & $32.01$ & $32.46$ & $32.98$ & $\textbf{33.24}$ \\
\\
\includegraphics[width=.145\textwidth]{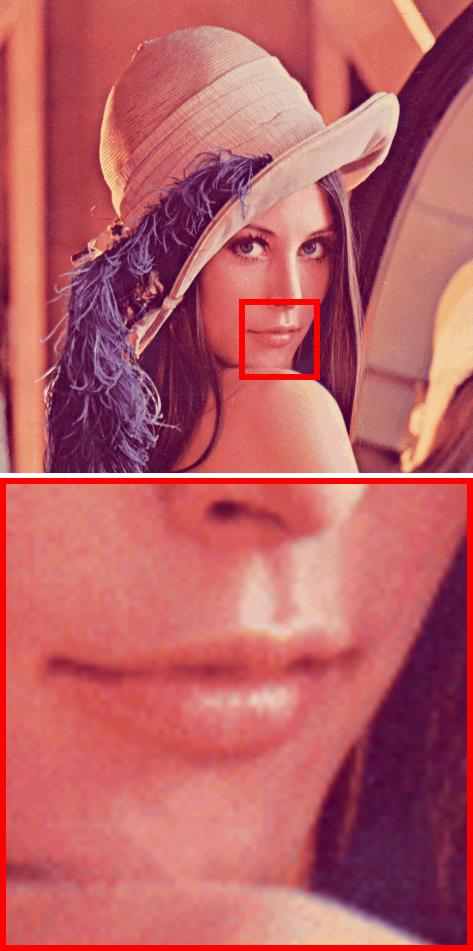} &
\includegraphics[width=.145\textwidth]{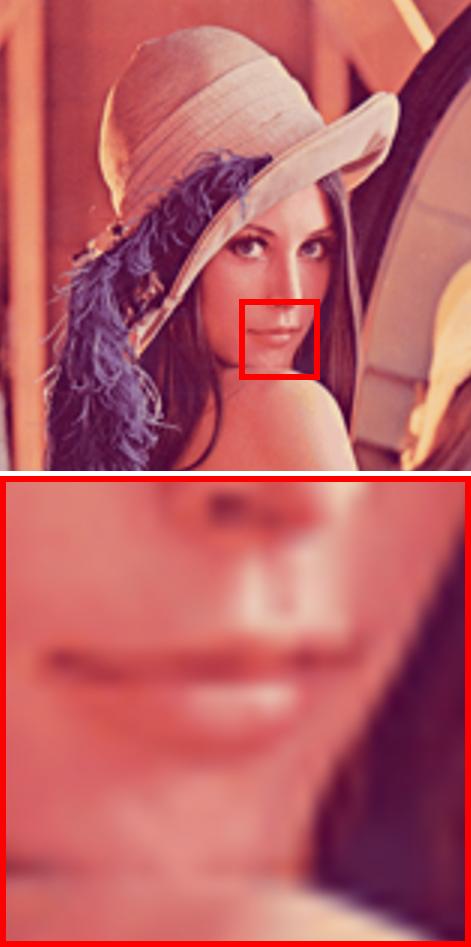} &
\includegraphics[width=.145\textwidth]{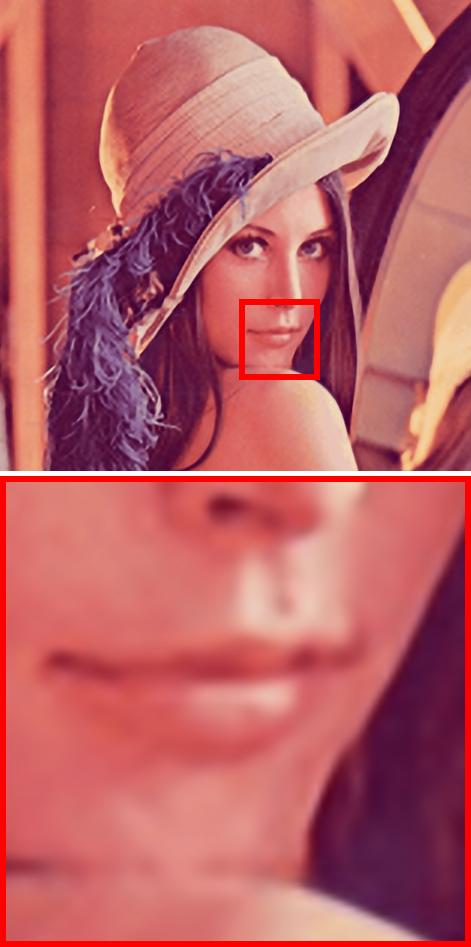} &
\includegraphics[width=.145\textwidth]{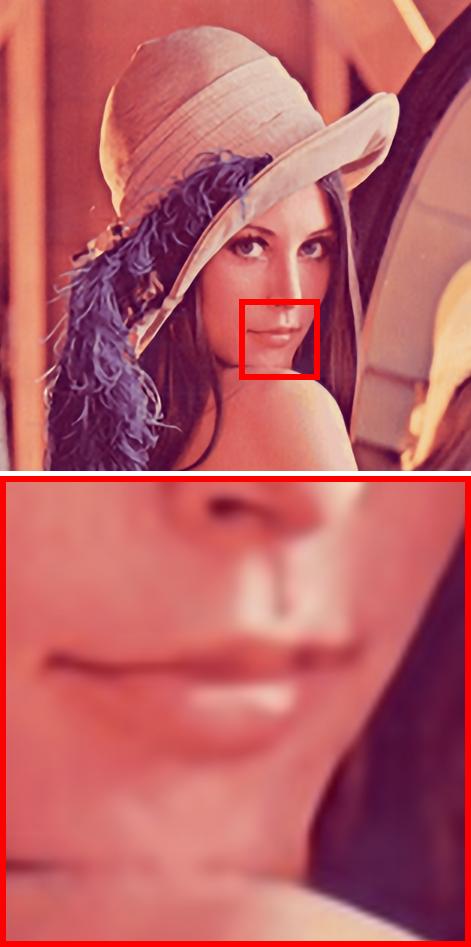} &
\includegraphics[width=.145\textwidth]{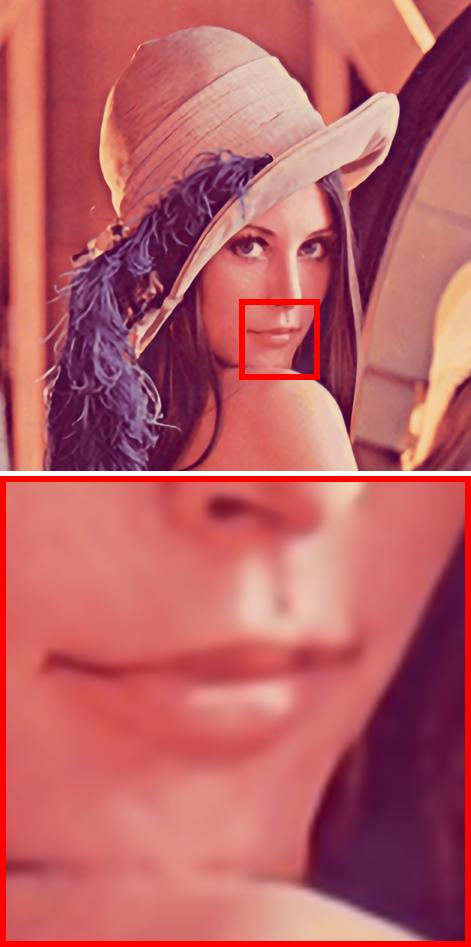} &
\includegraphics[width=.145\textwidth]{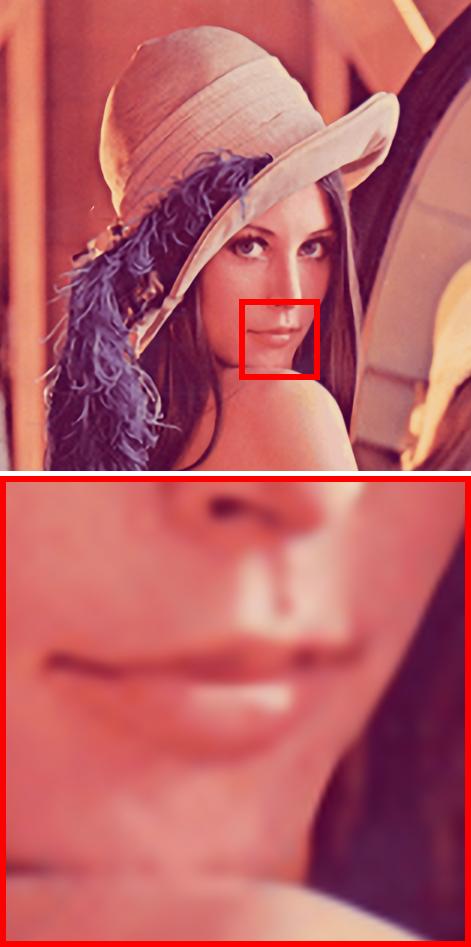} \\
 & $28.70$ & $31.09$ & $31.27$ & $31.45$ & $\textbf{31.67}$ \\
\\
\includegraphics[width=.145\textwidth]{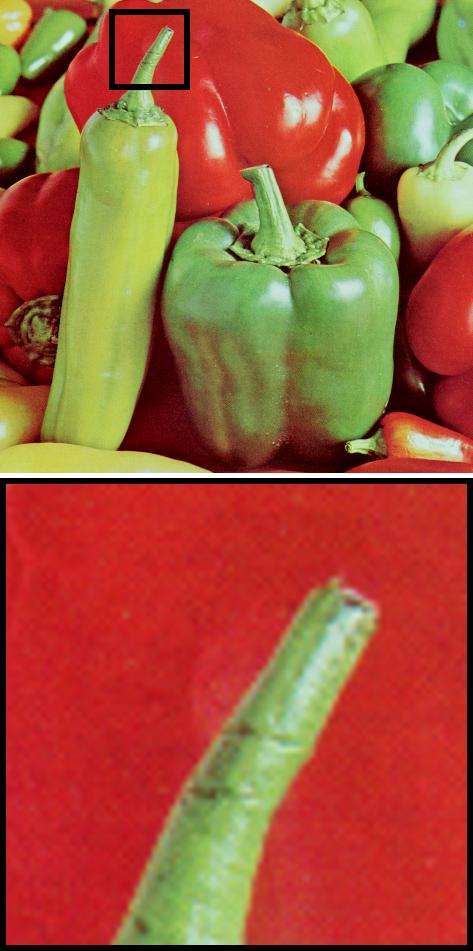} &
\includegraphics[width=.145\textwidth]{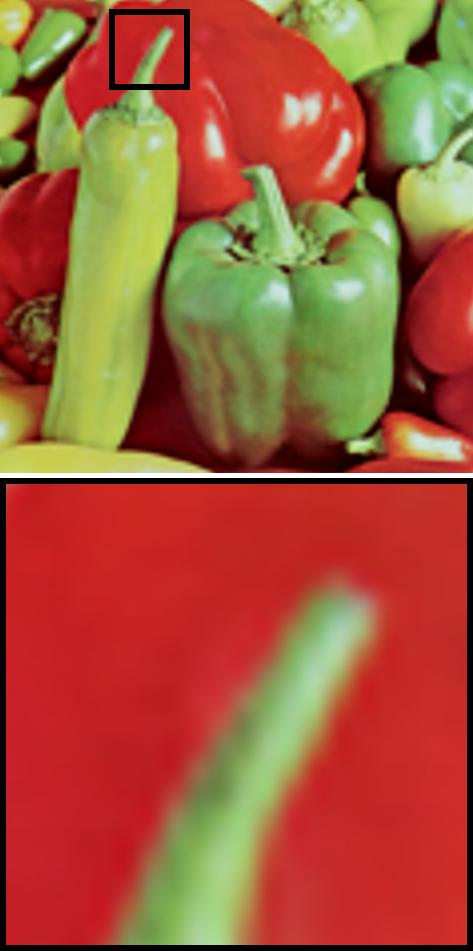} &
\includegraphics[width=.145\textwidth]{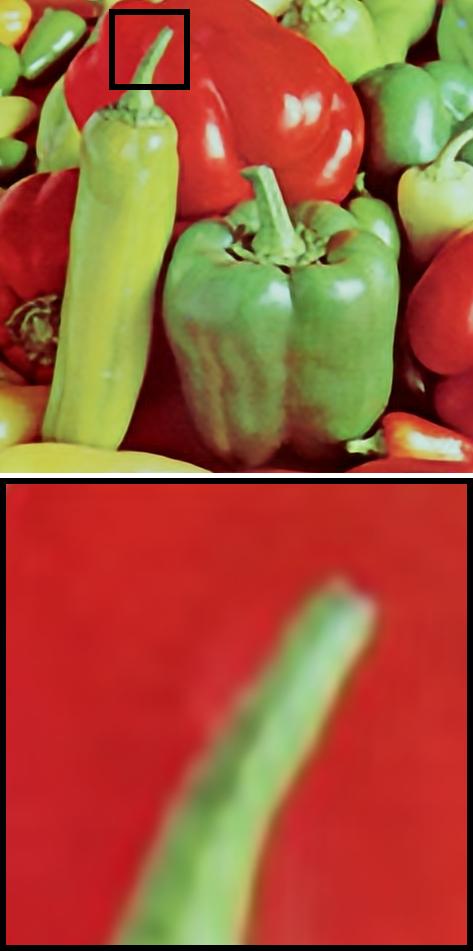} &
\includegraphics[width=.145\textwidth]{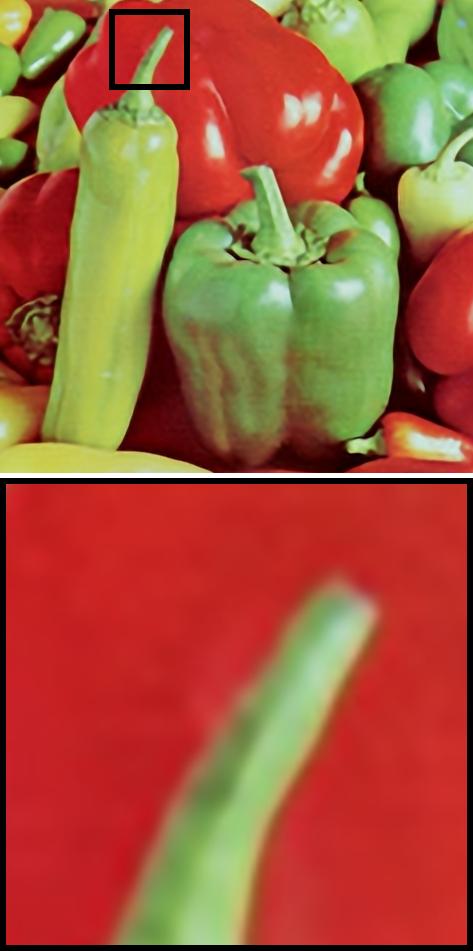} &
\includegraphics[width=.145\textwidth]{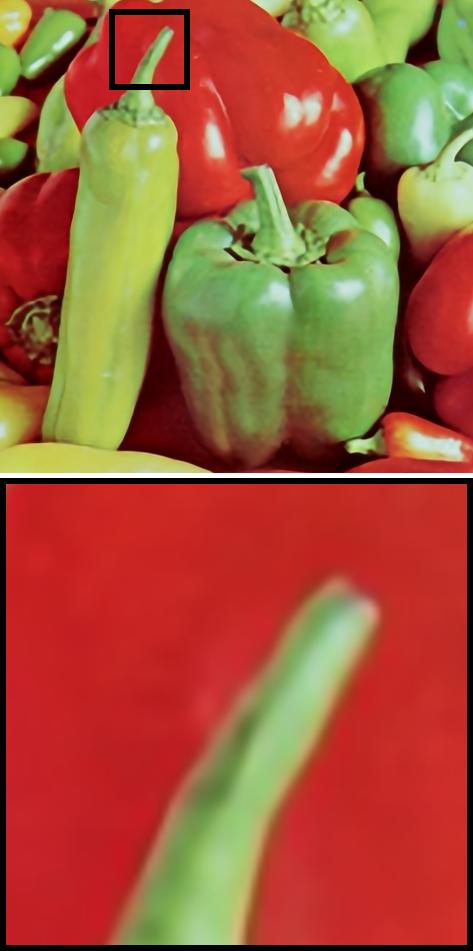} &
\includegraphics[width=.145\textwidth]{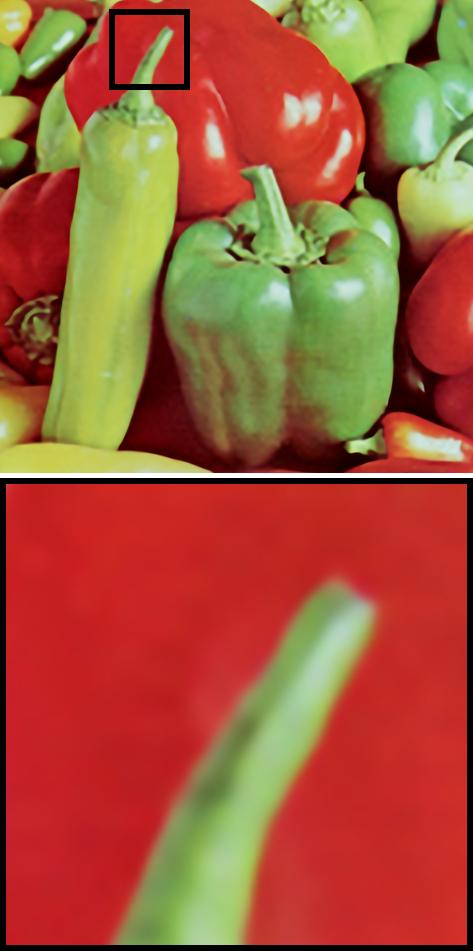} \\
 & $28.67$ & $29.98$ & $30.03$ & $30.31$ & $\textbf{30.96}$
\end{tabular}
\egroup
\caption{
Comparison of super-resolution for scale factor 2 (top row), scale factor 3 (middle row), and scale factor 4 (bottom row) with the corresponding PSNR ($dB$) scores.
}
\label{fig:superresolution}
\end{figure*}

The super-resolution problem is usually defined in absence of noise ($\sigma_d=0$), therefore we weight the prior by the inverse square root of the iteration number.
This policy starts with a rough regularization and reduces the prior weight in each iteration, leading to solutions that satisfy $\sigma_d=0$.
We compare our method to recent techniques by Kim et al.~\cite{Kim:2016:AIS} (SRCNN), Dong et al.~\cite{Dong:2016:ISR} (VDSR), and Zhang et al.~\cite{zhang2016beyond} (DnCNN-3), and TNRD by Chen and Pock~\cite{chen2016trainable}.
SRCNN, VDSR and DnCNN-3 train an end-to-end network by minimizing the $L_2$ loss between the output of the network and the high-resolution ground truth, and TNRD uses a learned reaction diffusion model.
While SRCNN and TNRD were trained separately for each scale, the VDSR and DnCNN-3 models were trained jointly on $\times 2, 3$ and $4$ (DnCNN-3 training included also denoising and JPEG artifact removal tasks).
For $\times 5$ super-resolution we used SRCNN and TNRD models that were trained on $\times 4$, and we used VDSR and DnCNN-3 models trained jointly on $\times 2, 3$ and $4$.
Tables~\ref{tbl:superResolutionSet5},~\ref{tbl:superResolutionSet14} compare the average PSNR of the super-resolved images from 'Set5' and 'Set14' datasets~\cite{Bevilacqua:2012:LCS,Zeyde:2010:SIS} for scale factors $\times 2, 3, 4,$ and $5$.
We compute PSNR values over cropped RGB images (where the crop size in pixels corresponds to the scale factor) for all methods. For SRCNN, however, we used a boundary of $13$ pixels to provide full support for their network.
While SRCNN, VDSR and DnCNN-3 solve directly for MMSE, our method solves for the MAP solution, which is not guaranteed to have better PSNR. Still, we achieve better results in average.
For scale factor $\times 5$ our method performs significantly better since our prior does not need to be trained for a specific scale.
Figure~\ref{fig:superresolution} shows visual comparisons to the super-resolution results from SRCNN~\cite{Dong:2016:ISR}, TNRD~\cite{chen2016trainable}, and DnCNN-3~\cite{zhang2016beyond} on three example images.
We exclude results of VDSR due to limited space and visual similarity with DnCNN-3.
Our natural image prior provides clean and sharp edges over all magnification factors.

\subsection{Non-Blind Deconvolution}

\begin{table}[t]
\begin{center}
\bgroup
\setlength{\tabcolsep}{3pt}
\begin{tabular}[c]{ccccc}
\hlineB{2}
  $ \sigma$ & Levin~\cite{levin2007image} & EPLL~\cite{Zoran:2011:LMN} & RTF-6~\cite{schmidt2016cascades}&  DAEP (Ours)\\
\hline
$ 2.55 $ & 31.09 & 32.51 & 32.51 & \textbf{32.69} \\
$ 7.65 $ & 27.40 & 28.42 & 21.44 & \textbf{28.95} \\
$ 12.75 $ & 25.36 & 26.13 & 16.03 & \textbf{26.87} \\
\hlineB{2}
\end{tabular}
\egroup
\end{center}
\caption{Average PSNR ($dB$) for non-blind deconvolution on Levin et al.'s~\cite{levin2007image} dataset for different noise levels.}
\label{tbl:deconvolution}
\end{table}

\begin{figure*}[t]
\centering
\bgroup
\setlength{\tabcolsep}{5pt}
\begin{tabular}{cccccc}
Ground Truth & Blurred & Levin et al.~\cite{levin2007image} & EPLL~\cite{Zoran:2011:LMN} & RTF-6~\cite{schmidt2016cascades} & DAEP (Ours) \\
\includegraphics[width=.145\textwidth]{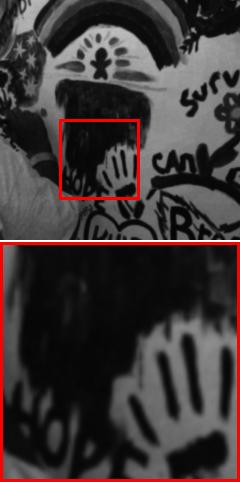} &
\includegraphics[width=.145\textwidth]{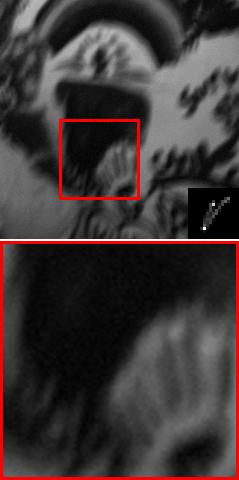} &
\includegraphics[width=.145\textwidth]{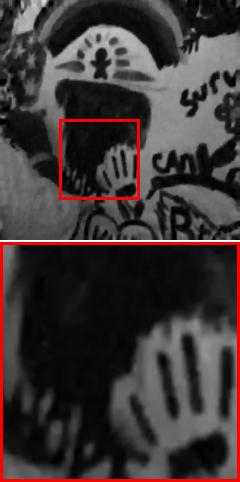} &
\includegraphics[width=.145\textwidth]{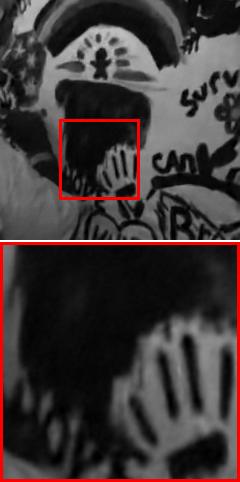} &
\includegraphics[width=.145\textwidth]{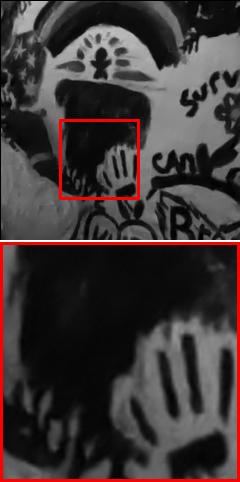} &
\includegraphics[width=.145\textwidth]{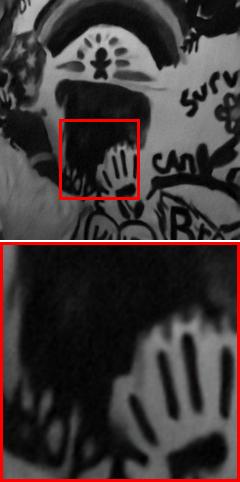} \\
 & $22.05$ & $30.88$ & $32.69$ & $32.82$ & $\textbf{33.64}$ \\
\\
\includegraphics[width=.145\textwidth]{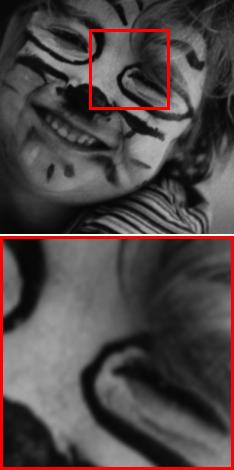} &
\includegraphics[width=.145\textwidth]{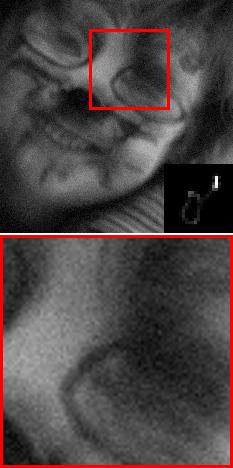} &
\includegraphics[width=.145\textwidth]{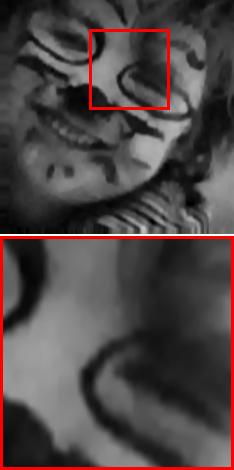} &
\includegraphics[width=.145\textwidth]{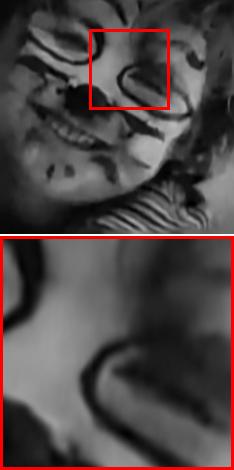} &
\includegraphics[width=.145\textwidth]{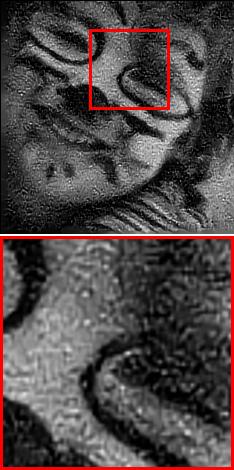} &
\includegraphics[width=.145\textwidth]{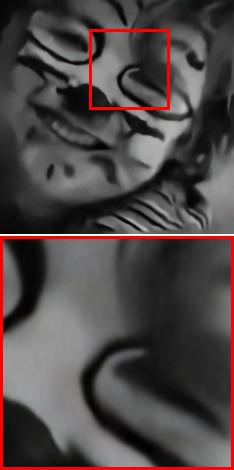} \\
 & $19.47$ & $28.22$ & $29.65$ & $21.82$ & $\textbf{30.68}$
\end{tabular}
\egroup
\caption{
Comparison of non-blind deconvolution with $\sigma = 2.55$ additive noise (top row) and $\sigma = 7.65$ additive noise (bottom row) with the corresponding PSNR ($dB$) scores.
The kernel is visualized in the bottom right of the blurred image.
}
\label{fig:deconvolution}
\end{figure*}

To evaluate and compare our method for non-blind deconvolution we used the dataset from Levin et al.~\cite{levin2007image} with four grayscale images and eight blur kernels in different sizes from $13 \times 13$ to $27 \times 27$.
We compare our results to Levin et al.~\cite{levin2007image}, Zoran and Weiss~\cite{Zoran:2011:LMN}, and Schmidt et al.~\cite{schmidt2016cascades} in Table~\ref{tbl:deconvolution}, where we show the average PSNR of the deconvolution for three levels of additive noise ($\sigma \in \{2.55, 7.65, 12.75 \}$).
Note that RTF-6~\cite{schmidt2016cascades} is only trained for noise level $\sigma=2.55$, therefore it does not perform well for other noise levels.
Figure~\ref{fig:deconvolution} provides visual comparisons for two deconvolution result images.
Our natural image prior achieves higher PSNR and produces sharper edges and less visual artifacts compared to Levin et al.~\cite{levin2007image}, Zoran and Weiss~\cite{Zoran:2011:LMN}, and Schmidt et al.~\cite{schmidt2016cascades}.

\begin{figure*}[t]
\centering
\small
\bgroup
\setlength{\tabcolsep}{5pt}
\begin{tabular}{ccccc}
\includegraphics[width=.18\textwidth]{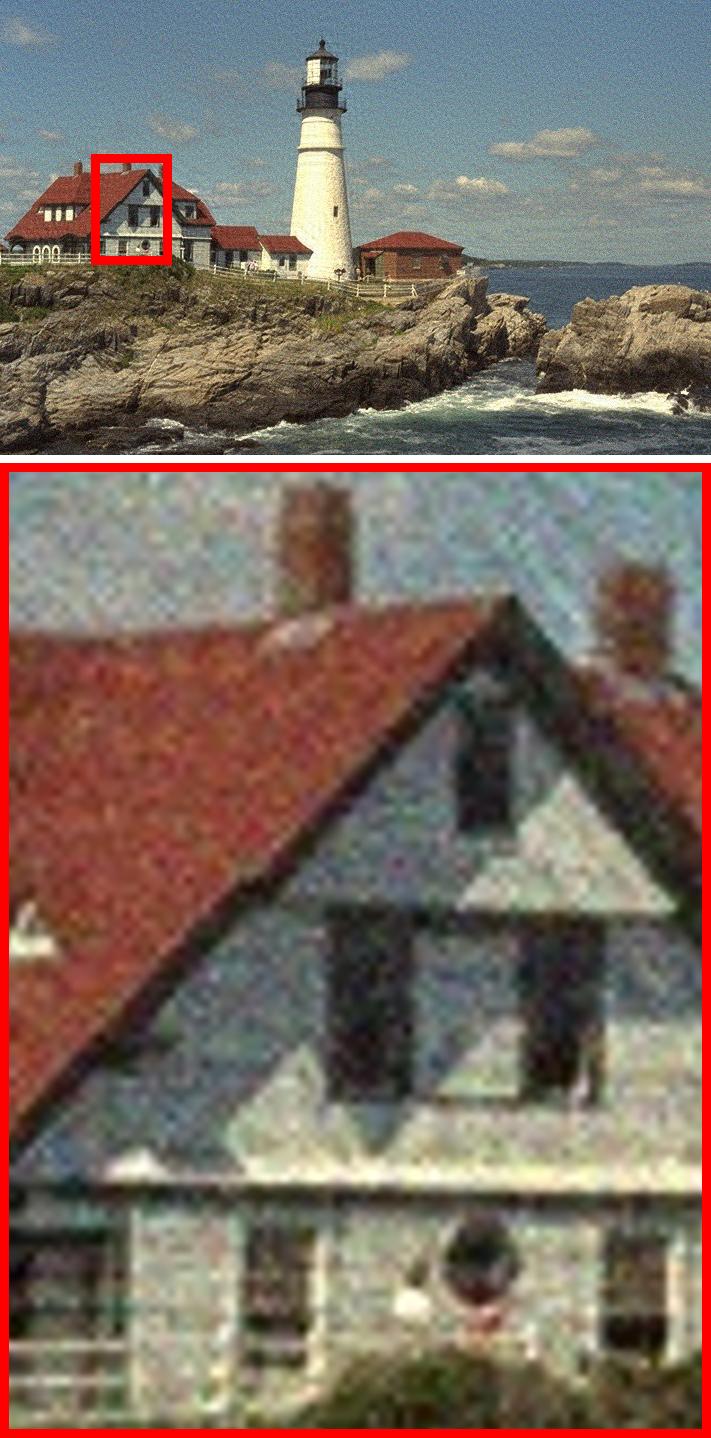} &
\includegraphics[width=.18\textwidth]{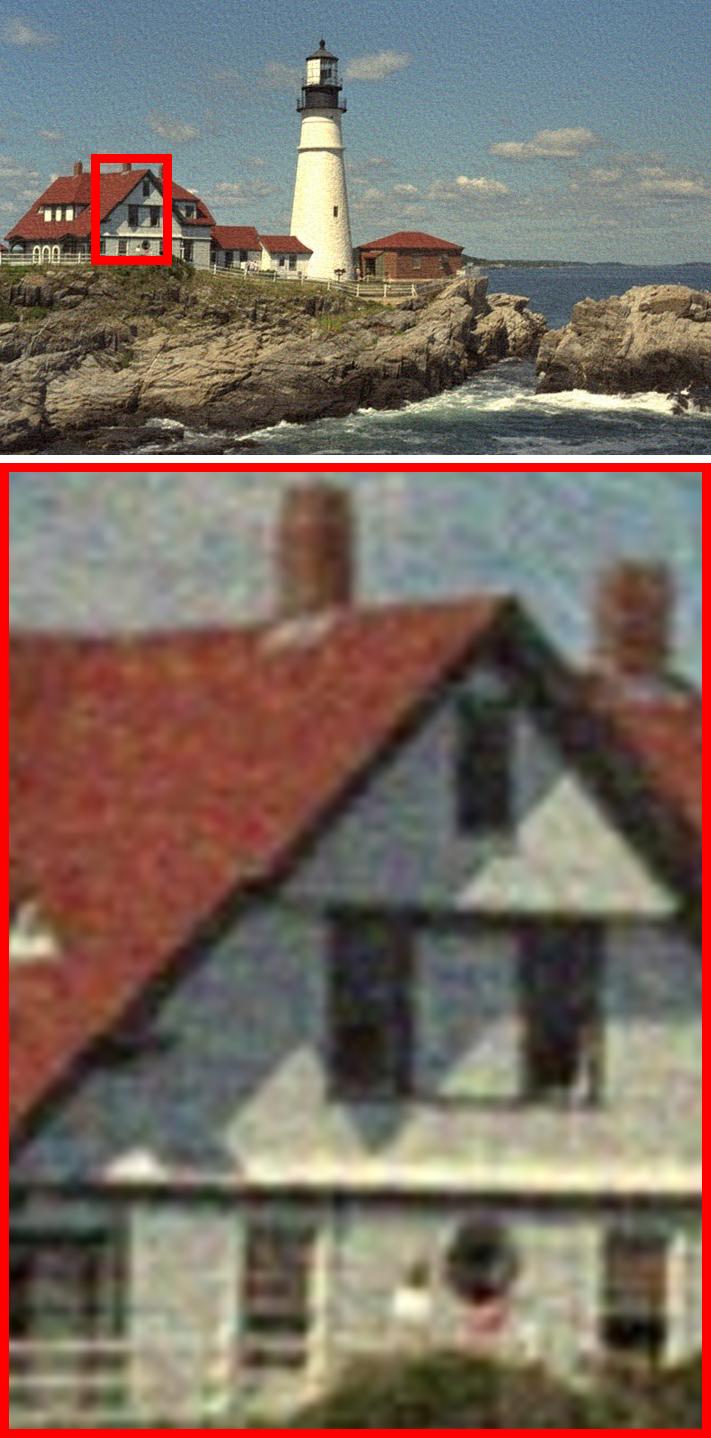} &
\includegraphics[width=.18\textwidth]{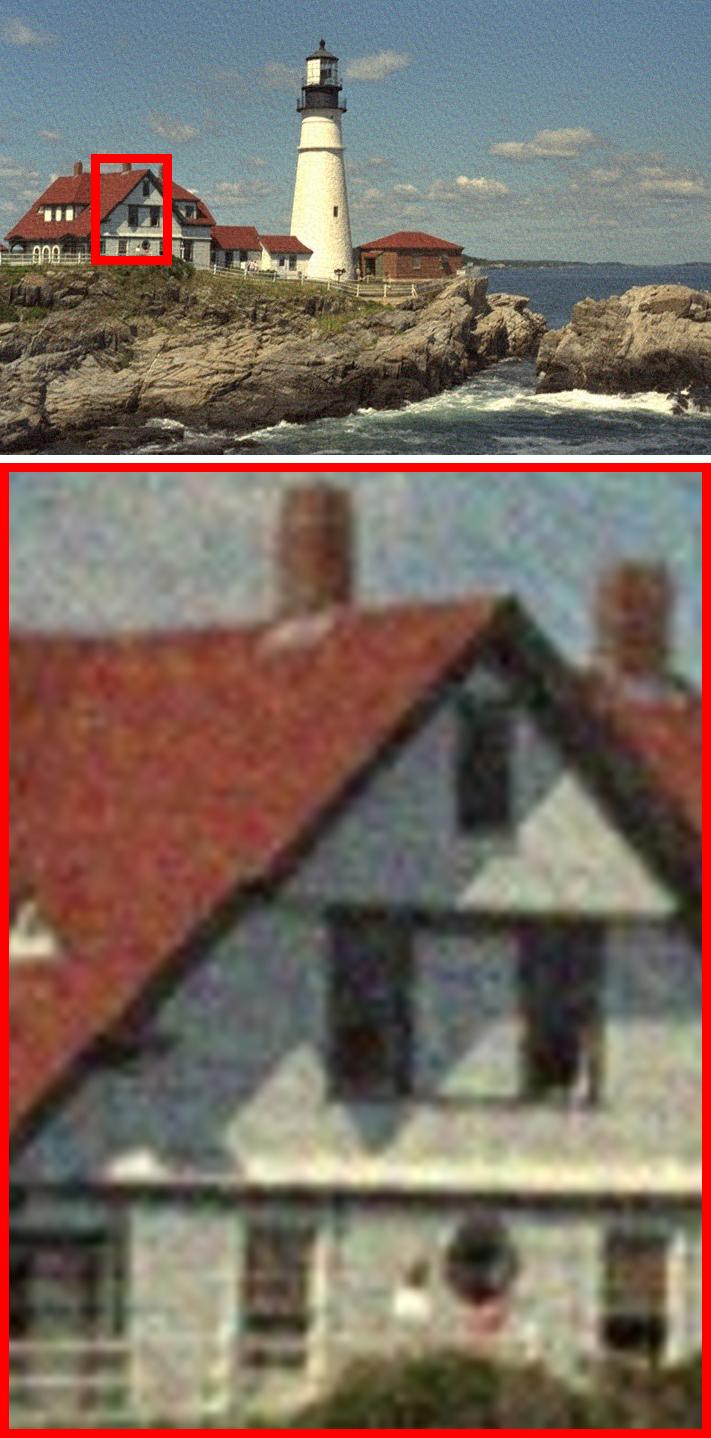} &
\includegraphics[width=.18\textwidth]{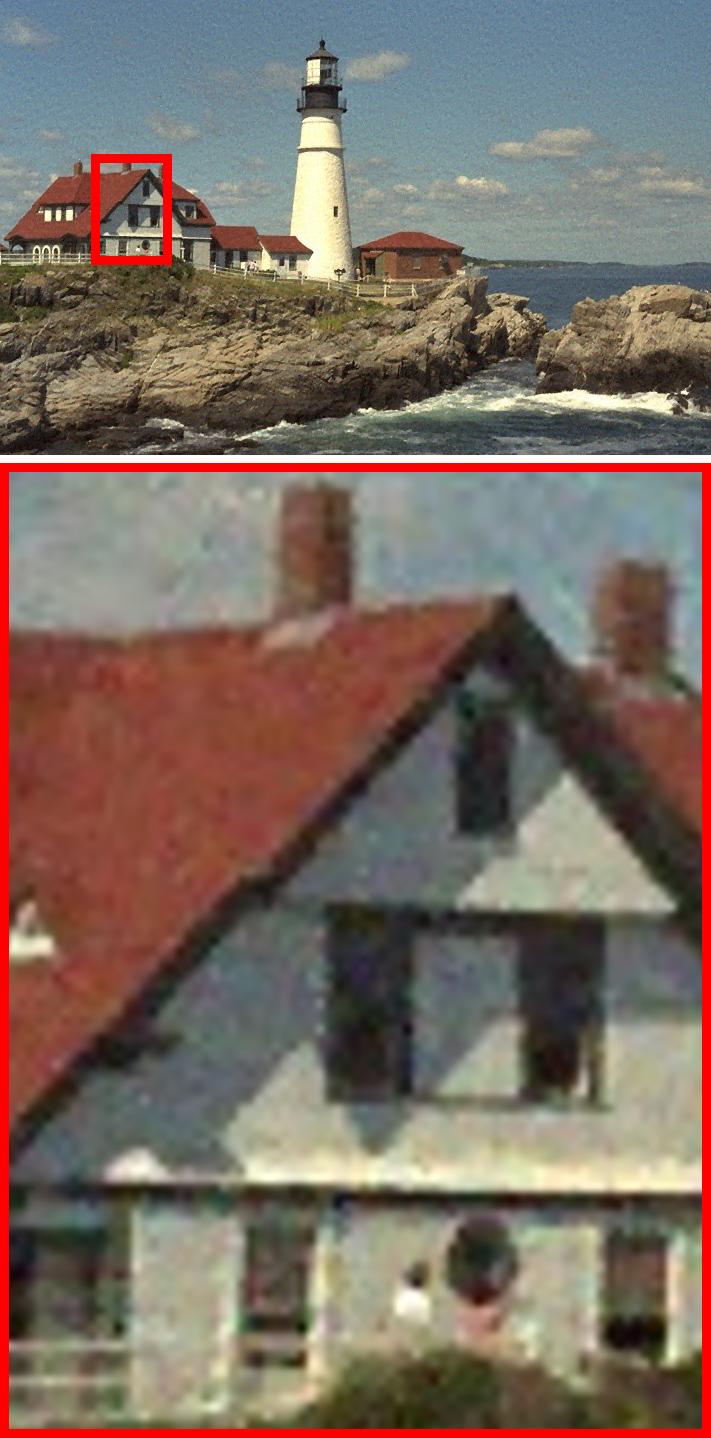} &
\includegraphics[width=.18\textwidth]{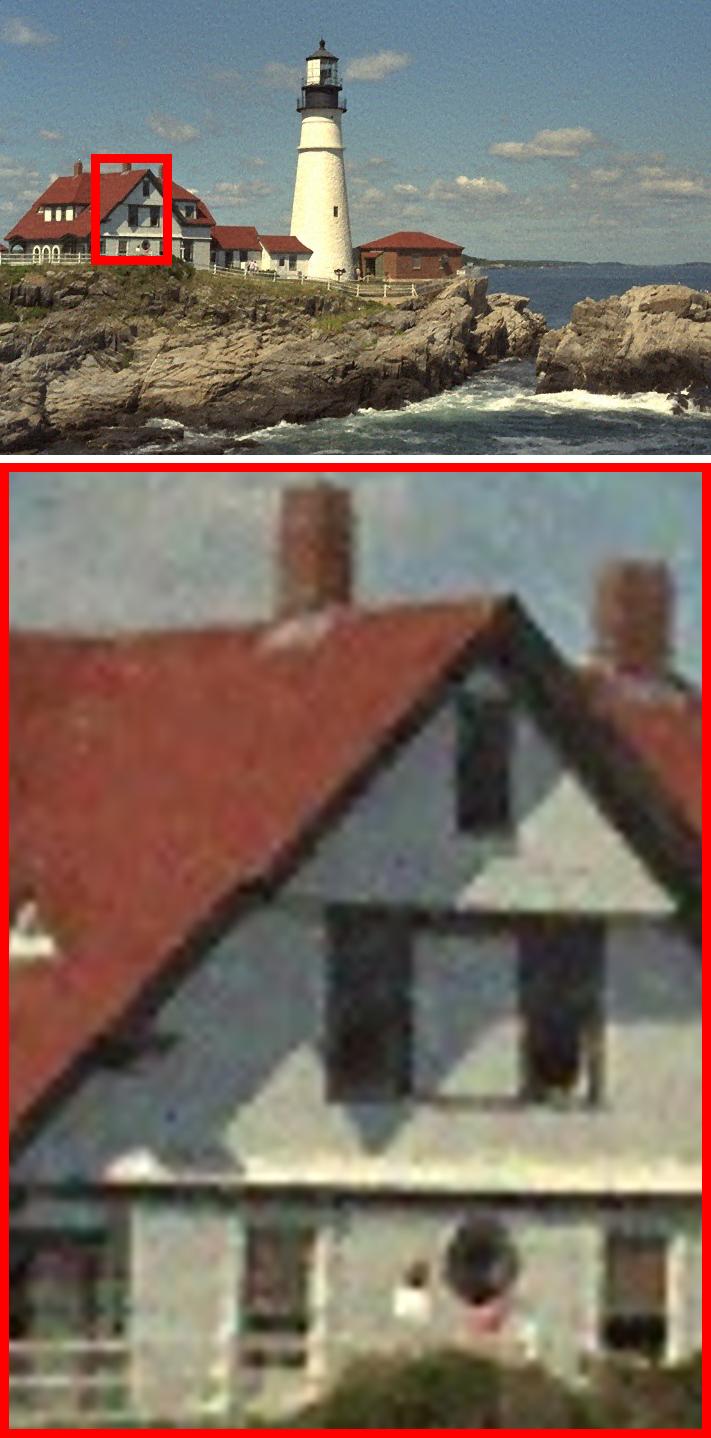} \\
Lucy Richardson & Zhou et al.~\cite{zhou2009good} & Levin et al.~\cite{levin2007image} ($L_2$) & Wang et al.~\cite{wang2008new} ($L_1$) & Wang et al.~\cite{wang2008new} (TV) \\
$24.38 / 24.47$ & $27.38 / 27.68$ & $27.04 / 27.37$ & $27.68 / 28.23$ & $28.63 / 29.25$ \\
\\
\includegraphics[width=.18\textwidth]{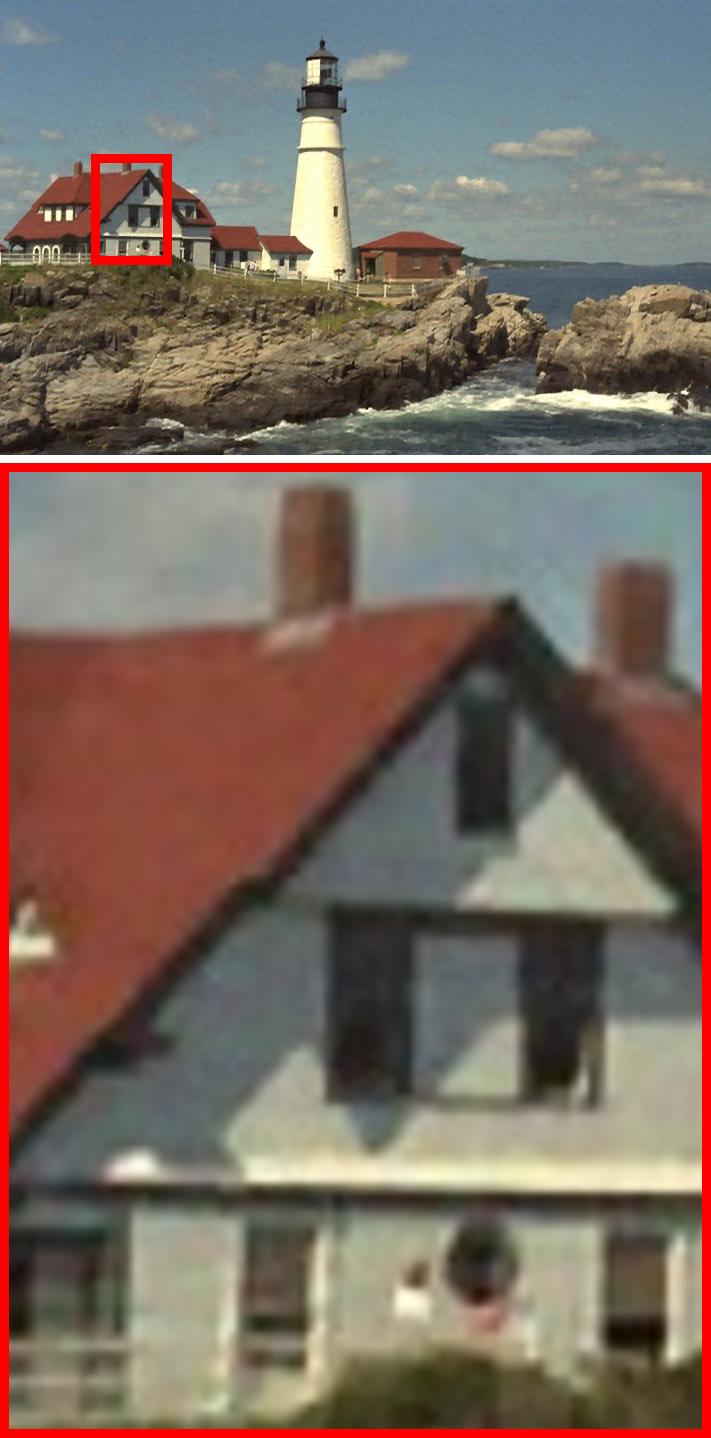} &
\includegraphics[width=.18\textwidth]{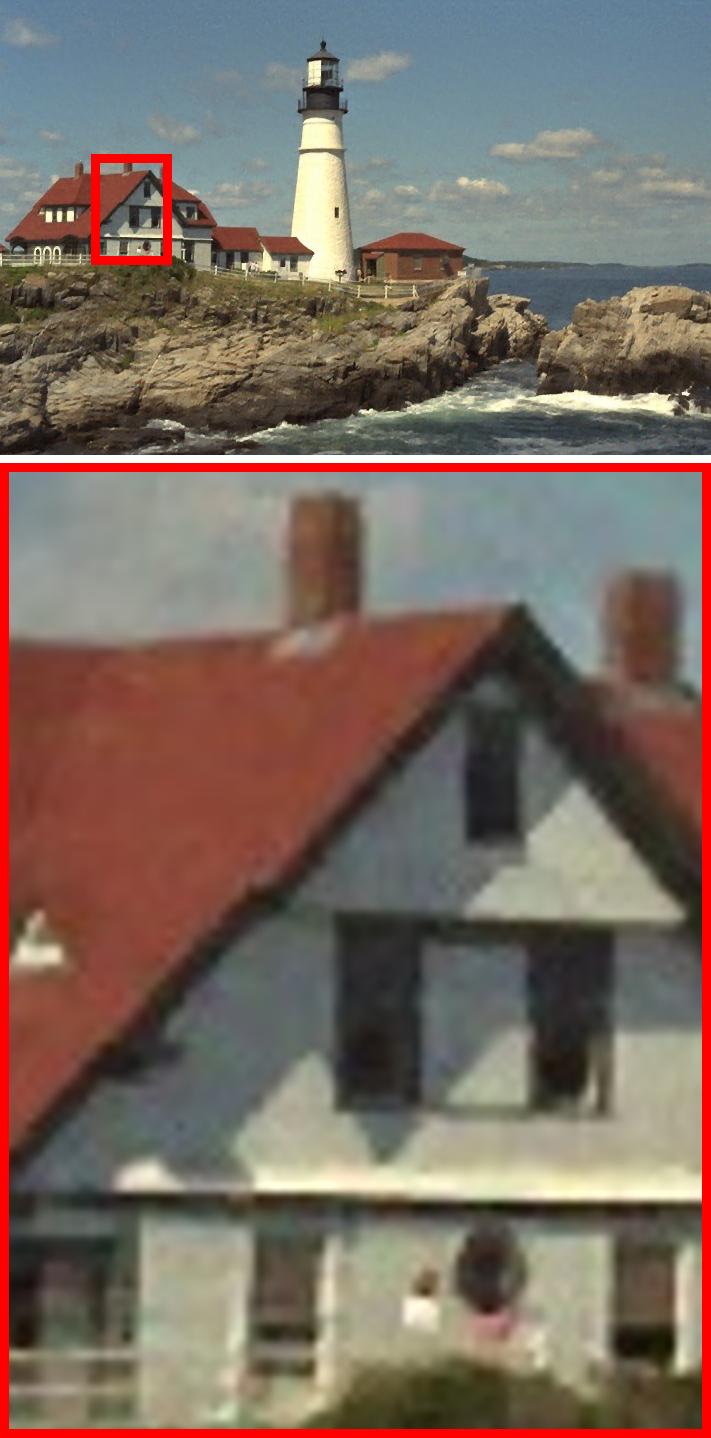} &
\includegraphics[width=.18\textwidth]{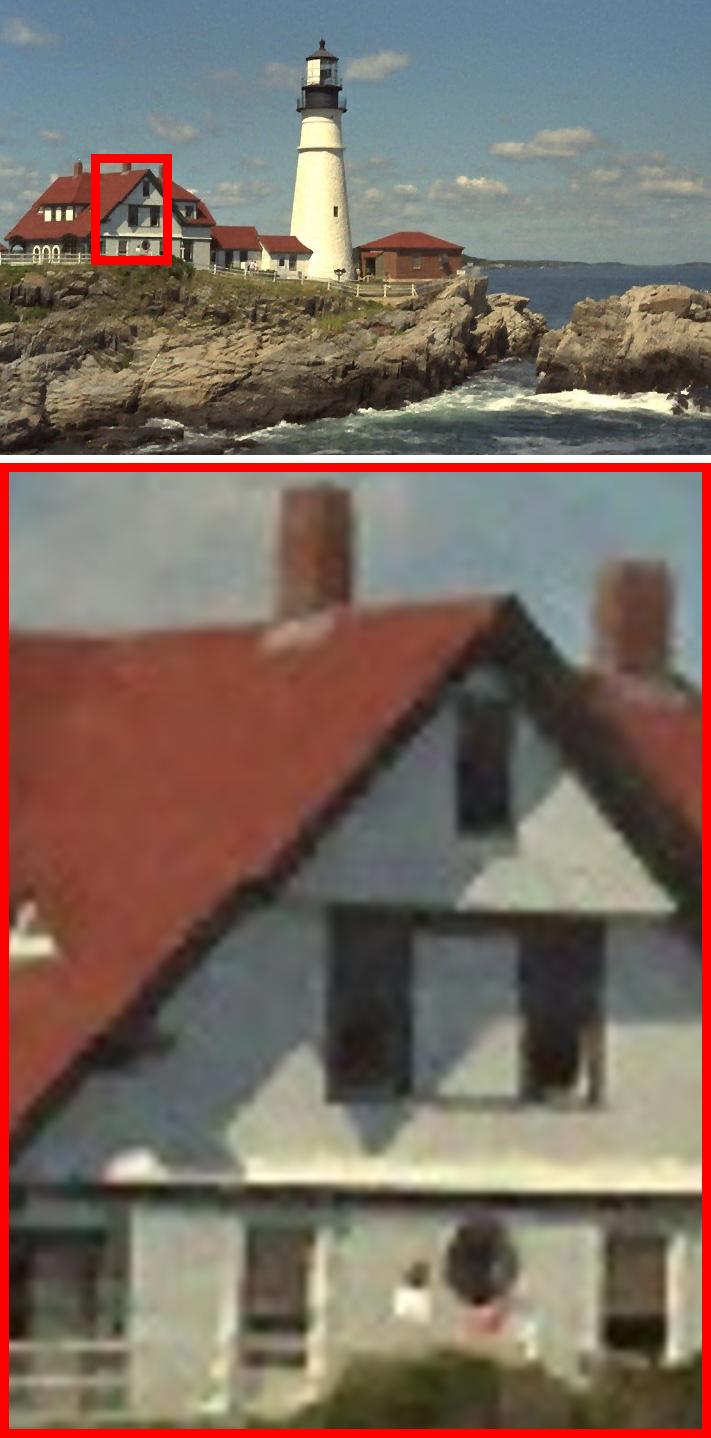} &
\includegraphics[width=.18\textwidth]{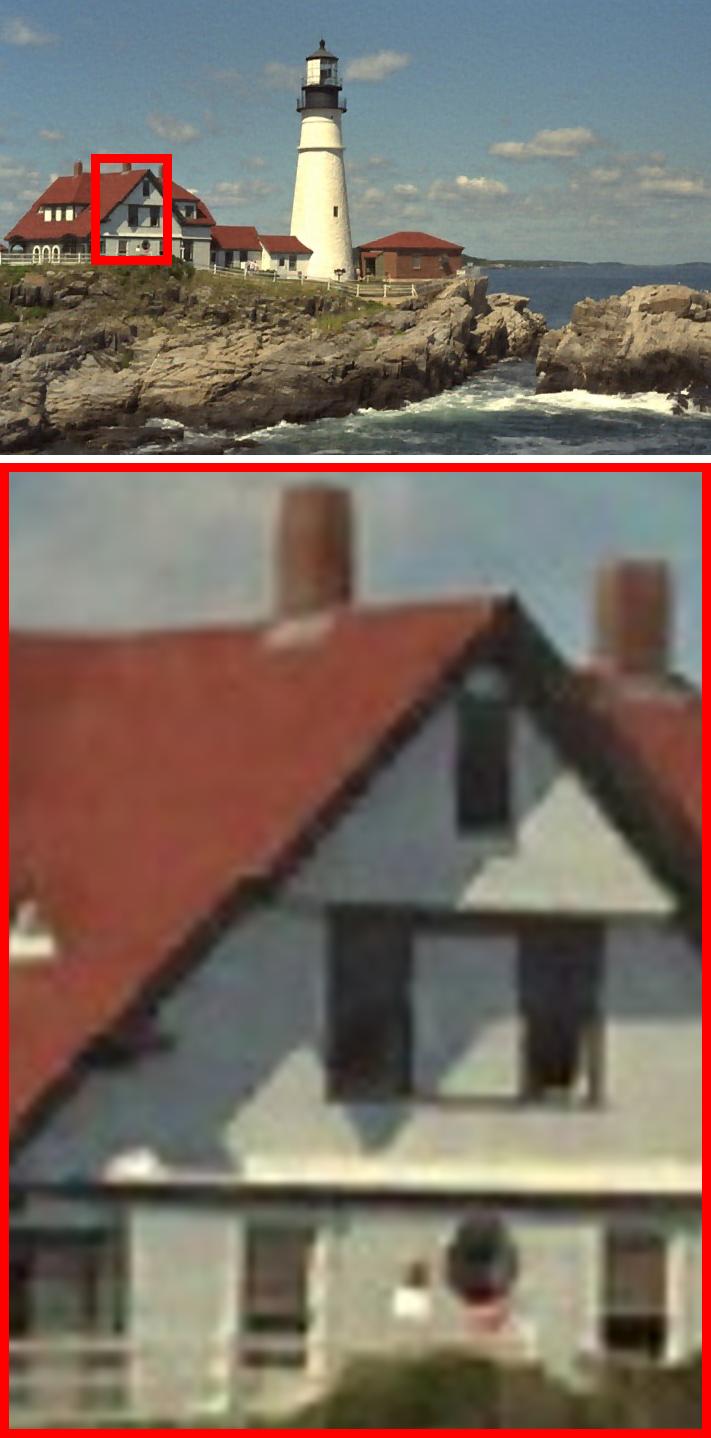} &
\includegraphics[width=.18\textwidth]{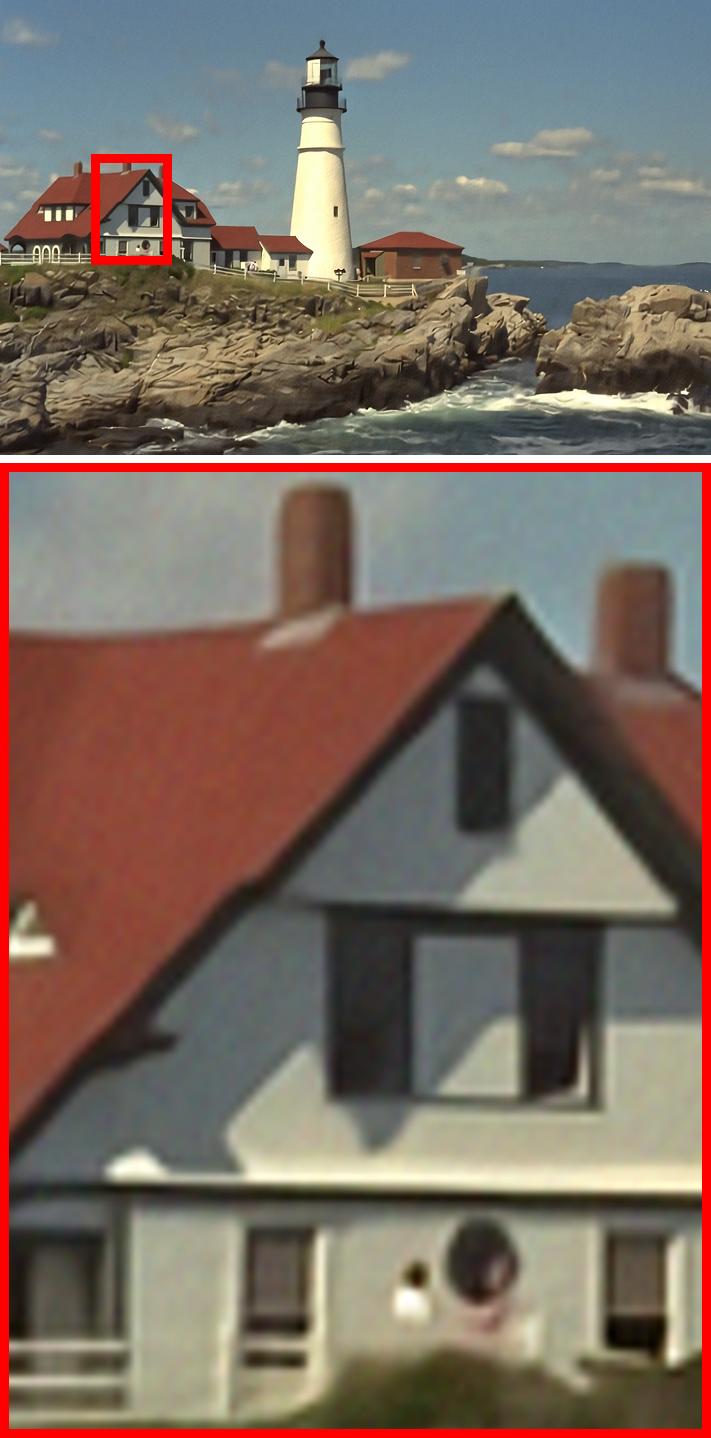} \\
Levin et al.~\cite{levin2007image} (IRLS) & Shan et al.~\cite{shan2008high} & Krishnan, Fergus~\cite{krishnan2009fast} & Fortunato, Oliveira~\cite{FortunaoOliveira2014FD} & DAEP (Ours) \\
$28.96 / 30.15$ & $28.97 / 30.01$ & $29.15 / 30.18$ & $29.25 / 30.34$ & $\textbf{29.92} / \textbf{31.07}$
\end{tabular}
\egroup
\caption
{Comparison of non-blind deconvolution methods on the 21st image from the Kodak image set~\cite{KodakSuite}. For each method, we report the PSNR ($dB$) of the visualized image (left) and the average PSNR on the whole set (right). The results of other methods were reproduced from Fortunato and Oliveira~\cite{FortunaoOliveira2014FD} for ease of comparison.}
\label{fig:deconvolutionColor}
\end{figure*}

We performed an additional comparison on color images similar to Fortunato and Oliveira~\cite{FortunaoOliveira2014FD} using $24$ color images from the Kodak Lossless True Color Image Suite from PhotoCD PCD0992~\cite{KodakSuite}.
The images are blurred with a $19 \times 19$ blur kernel from Krishnan and Fergus~\cite{krishnan2009fast} and $1\%$ noise is added.
Figure~\ref{fig:deconvolutionColor} shows visual comparisons and average PSNRs over the whole dataset.
Our method produces much sharper results and achieves a higher PSNR in average over this dataset.

\subsection{Discussion}

\begin{figure*}
\centering
\bgroup
\setlength{\tabcolsep}{5pt}
\begin{tabular}{cccc}
Masked $70\%$ of Pixels & Our Reconstruction & Input with $10\%$ Noise & Our Reconstruction \\
$6.13dB$ & $30.68dB$ & $20.47dB$ & $31.05dB$ \\
\includegraphics[width=.23\textwidth]{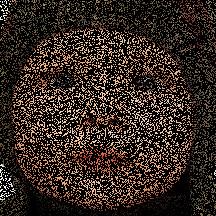} &
\includegraphics[width=.23\textwidth]{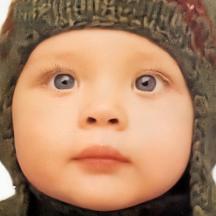} &
\includegraphics[width=.23\textwidth]{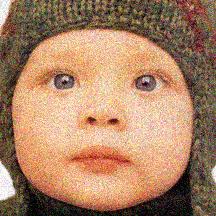} &
\includegraphics[width=.23\textwidth]{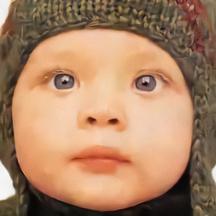} 
\end{tabular}
\egroup
\caption
{Restoration of images corrupted by noise and holes using the same autoencoding prior as in our other experiments.}
\label{fig:otherDeg}
\end{figure*}

A disadvantage of our approach is that it requires the solution of an optimization problem to restore each image. In contrast, end-to-end trained networks perform image restoration in a single feed-forward pass. For the increase in runtime computation, however, we gain much flexibility. With a single autoencoding prior, we obtain not only state of the art results for non-blind deblurring with arbitrary blur kernels and super-resolution with different magnification factors, but also successfully restore images corrupted by noise or holes as shown in Figure~\ref{fig:otherDeg}. 

Our approach requires some user defined parameters (mean shift kernel size $\sigma_{\eta}$ for DAE training and restoration, weight of the prior $\gamma$). While we use the same parameters for all experiments reported here, other applications may require to adjust these parameters. For example, we have experimented with image denoising (Figure~\ref{fig:otherDeg}), but so far we have not achieved state of the art results. 
We believe that this may require an adaptive kernel width for the DAE, and further fine-tuning of our parameters.

\section{Conclusions}

We introduced a natural image prior based on denoising autoencoders (DAEs). Our key observation is that optimally trained DAEs provide mean shift vectors on the true data density. Our prior minimizes the distances of restored images to their local means (the length of their mean shift vectors). This is powerful since mean shift vectors vanish at local extrema of the true density smoothed by the mean shift kernel. Our results demonstrate that a single DAE prior achieves state of the art results for non-blind image deblurring with arbitrary blur kernels and image super-resolution at different magnification factors. In the future, we plan to apply our autoencoding priors to further image restoration problems including denoising, colorization, or non-uniform and blind deblurring.
While we used Gaussian noise to train our autoencoder, it is possible to use other types of data degradation for DAE training. 
Hence, we will investigate other DAE degradations to learn different data representations or use a mixture of DAEs for the prior.


{\small
\bibliographystyle{ieee}
\bibliography{egbib}
}

\end{document}